\documentclass[10pt,twocolumn,letterpaper]{article}

\makeatletter
\@namedef{ver@everyshi.sty}{}
\makeatother
\usepackage{tikz}
\usetikzlibrary{positioning, fit}

\usepackage[pagenumbers]{cvpr} 

\usepackage{url}
\usepackage{multirow}
\usepackage{booktabs}
\usepackage{pifont}
\usepackage{graphicx}
\usepackage{amsmath}
\usepackage{algorithm}
\usepackage{mathtools}
\usepackage{algorithm}
\usepackage{algpseudocode}
\usepackage{listings}
\usepackage{wrapfig}
\usepackage{subcaption}
\usepackage{colortbl}
\usepackage{array}
\usepackage{float}
\usepackage{xcolor}
\usepackage{capt-of}
\usepackage{lipsum}
\usepackage[none]{hyphenat}
\usetikzlibrary{calc, positioning}

\usepackage{float}  
\usepackage{placeins}

\newcommand{\paragraphVspace}

\newcommand{\cmark}{\color{green}\ding{51}}%
\newcommand{\xmark}{\color{red}\ding{55}}%

\newif\ifdrafting
\draftingtrue 
\ifdrafting
    \newcommand{\todo}[1]{{\leavevmode\color[rgb]{0,0,1}[TODO: #1]}}

    \newcommand{\zh}[1]{{\leavevmode\color[rgb]{1,0.5,1}[Zehao: #1]}}

\else
    \newcommand{\todo}[1]{}
    \newcommand{\zh}[1]{}
\fi

\newcommand{\topic}[1]{\noindent\textbf{#1}}
\newcommand{\secref}[1]{Sec.~\ref{sec:#1}}
\newcommand{\figref}[1]{Fig.~\ref{fig:#1}}
\newcommand{\tabref}[1]{Tab.~\ref{tab:#1}}

\definecolor{cvprblue}{rgb}{0.21,0.49,0.74}
\usepackage[pagebackref,breaklinks,colorlinks,allcolors=cvprblue]{hyperref}

\def\paperID{3846} 
\def\confName{CVPR}
\def\confYear{2025}

\title{SceneCrafter: Controllable Multi-View Driving Scene Editing} 

\author{
Zehao Zhu\textsuperscript{1, 2},
Yuliang Zou\textsuperscript{1},
Chiyu Max Jiang\textsuperscript{1},
Bo Sun\textsuperscript{1},
Vincent Casser\textsuperscript{1},
Xiukun Huang\textsuperscript{1}\\
Jiahao Wang\textsuperscript{3},
Zhenpei Yang\textsuperscript{1},
Ruiqi Gao\textsuperscript{4},
Leonidas Guibas\textsuperscript{4},
Mingxing Tan\textsuperscript{1},
Dragomir Anguelov\textsuperscript{1}\\
\textsuperscript{1}Waymo, \textsuperscript{2}University of Texas at Austin, \textsuperscript{3}Johns Hopkins University, \textsuperscript{4}Google DeepMind
}

\begin{document}

\twocolumn[{
\renewcommand\twocolumn[1][]{#1}
\maketitle
\vspace{-3em}
\begin{center}
\centering
\includegraphics[width=0.94\linewidth]{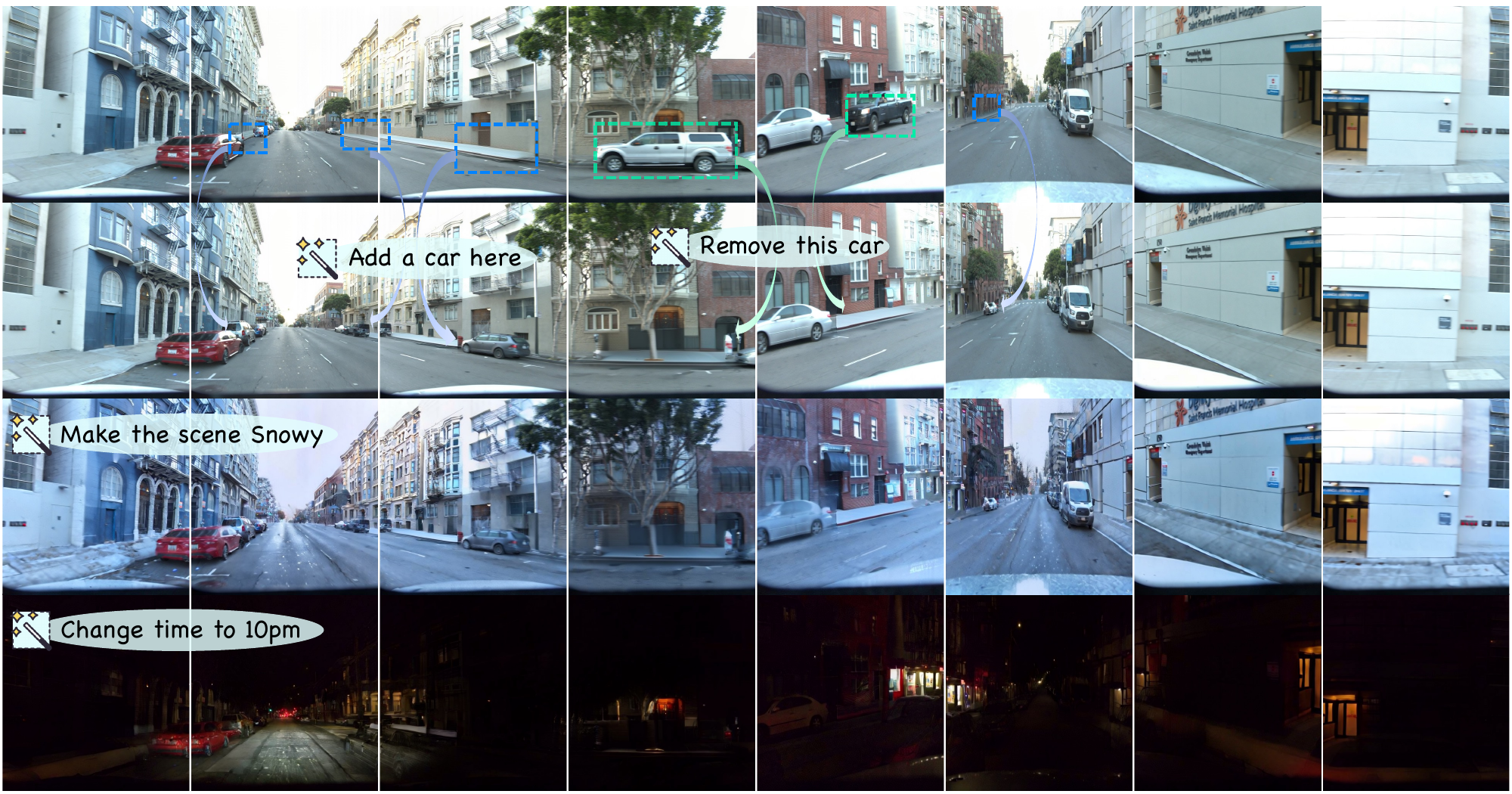}
\vspace{-0.8em}
\captionof{figure}{SceneCrafter is a versatile and dexterous editor for realistic 3D-consistent manipulation of driving scenes captured from multiple camera angles. It allows users to seamlessly insert or remove arbitrary objects in the foreground (second row) and modify global features like weather (third row) and time of day (fourth row), while preserving fine-grained details of scene layout and geometry.}
\label{fig:teaser}
\end{center}
}]

\begin{abstract}
Simulation is crucial for developing and evaluating autonomous vehicle (AV) systems. Recent literature builds on a new generation of generative models to synthesize highly realistic images for full-stack simulation. However, purely synthetically generated scenes are not grounded in reality and have difficulty in inspiring confidence in the relevance of its outcomes. Editing models, on the other hand, leverage source scenes from real driving logs, and enable the simulation of different traffic layouts, behaviors, and operating conditions such as weather and time of day. While image editing is an established topic in computer vision, it presents fresh sets of challenges in driving simulation: (1) the need for cross-camera 3D consistency, (2) learning ``empty street" priors from driving data with foreground occlusions, and (3) obtaining paired image tuples of varied editing conditions while preserving consistent layout and geometry. 
To address these challenges, we propose SceneCrafter, a versatile editor for realistic 3D-consistent manipulation of driving scenes captured from multiple cameras. We build on recent advancements in multi-view diffusion models, using a fully controllable framework that scales seamlessly to multi-modality conditions like weather, time of day, agent boxes and high-definition maps. To generate paired data for supervising the editing model, we propose a novel framework on top of Prompt-to-Prompt~\cite{hertz2022prompt} to generate geometrically consistent synthetic paired data with global edits. We also introduce an alpha-blending framework to synthesize data with local edits, leveraging a model trained on empty street priors through novel masked training and multi-view repaint paradigm. 
SceneCrafter demonstrates powerful editing capabilities and achieves state-of-the-art realism, controllability, 3D consistency, and scene editing quality compared to existing baselines.
\end{abstract}
\section{Introduction}
\label{sec:intro}
\begin{figure*}[t]
\centering
\includegraphics[width=0.98\textwidth]{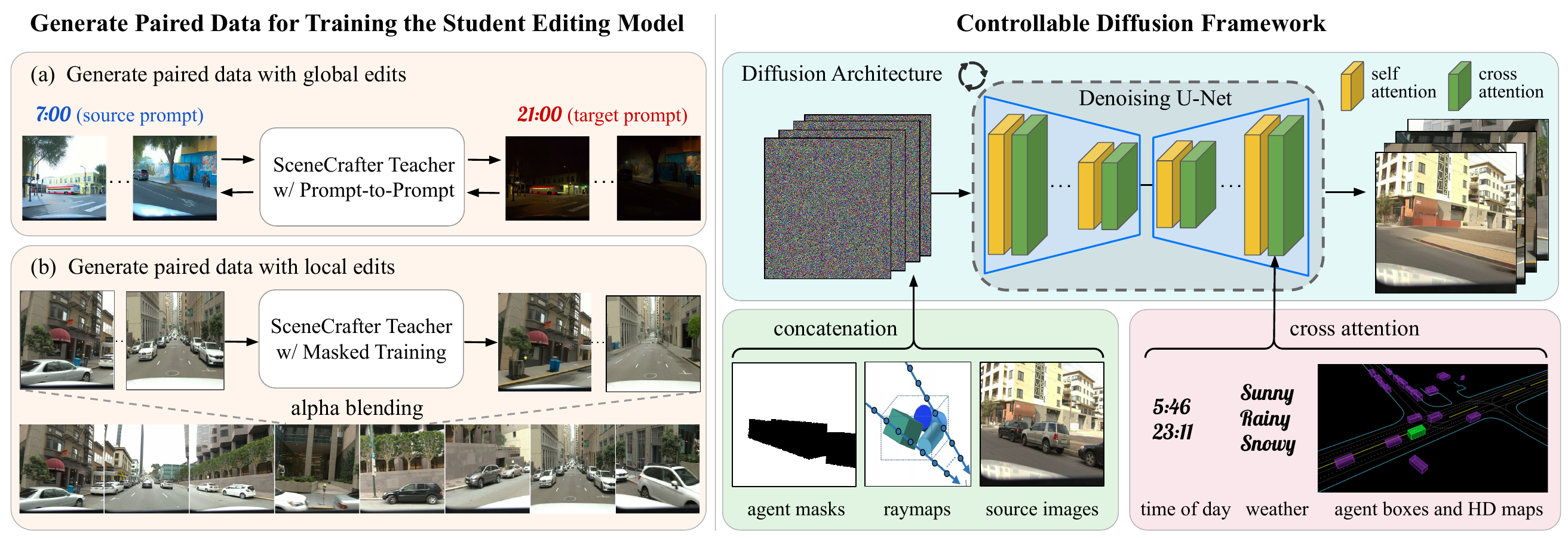}
\vspace{-3mm}
\caption{
\textbf{Overview.}
Our method consist of two main stages.
First, we train two teacher models to synthesize a large-scale paired dataset with several novel ideas (\secref{generate_data}).
We then train a unified student model with the generated data for 3D-consistent scene editing (\secref{edit}).
}
\label{fig:method}
\end{figure*}

Simulation is a core component in autonomous vehicle (AV) development for enabling system-level evaluation and quality hillclimbing at lower cost and faster turnaround times than real-world testing. Critical use-cases are evaluating the performance of the AV in new traffic scenarios (different agent location and behavior) as well as new operating conditions (such as different weather).

Most existing works in the driving simulation literature reduce simulation to an agent behavior simulation problem~\cite{caesar2021nuplan,montali2024waymo,gulino2024waymax} so as to enable closed-loop simulation of motion planners. This formulation bypasses the perception system, which limits their applicability and can cause simulations to be untrustworthy under behavioral changes due to pose divergence. These methods are unable to simulate changing operating conditions challenging the perception system, nor can they support evaluation of end-to-end planners~\cite{hu2023planning,hwang2024emmaendtoendmultimodalmodel} requiring realistic sensor inputs.

Sensor simulation approaches have benefited from recent advances in reconstructive modeling through neural fields/primitives~\cite{mildenhall2020nerf,kerbl20233d} or image/video generation models \cite{sohl2015deep,song2019generative}. While reconstructive techniques are faithful to real scenes, they lack the flexibility of efficiently simulating varied lighting and weather conditions, or manipulating existing objects. Generative sensor simulation methods are typically formulated as an image or video generation task conditioned on attributes such as text descriptions or scene layout. However, such scenes are usually not grounded in real scenes and do not inspire confidence in their actual occurrence under real driving conditions. Editing models, on the other hand, offer the best combination of realism-grounding as well as the ability to rely upon large data priors for flexible edits.

While there is a large body of work in generic image editing tasks, there is limited work, to our knowledge, in applying this in the context of multi-view driving imagery. In particular, to facilitate the two aforementioned sensor simulation tasks of evaluating under new traffic scenarios and operating conditions, we require two types of editing modalities: local foreground editing such as agent removal and injection conditioned on agent location, size and type, as well as global editing pertaining different operating conditions, such as varying weather conditions and time of day. To train the final editing model, we require paired image tuples before and after the edit to supervise the final editing model. Obtaining paired supervision data for local and global editing poses different sets of challenges.

Local editing requires pairs of ``empty streets" (with no agents) and ``populated streets". However as we show in \secref{ablation}, directly applying inpainting techniques such as RePaint \cite{lugmayr2022repaint} to erase agents does not result in high quality results due to the model not having good priors of ``empty streets" as almost all training data contains ``populated streets". We develop a novel training paradigm that we coin \emph{masked training} that enables us to learn to priors of empty streets from datasets of populated streets, resulting in high quality ``empty street" priors.
It enables us to generate high quality pairs of populated and unpopulated scenes, from which we can selectively curate pairs of partially populated scenes via alpha blending to train the editing model.

Global editing poses a different set of challenges. While seminal work such as Prompt-to-Prompt \cite{hertz2022prompt} and InstructPix2Pix \cite{brooks2022instructpix2pix} offer an exciting new avenue for creating paired synthetic images, directly applying Prompt-to-Prompt to the multi-camera
setting does not perform well due to the absence of text conditions whose attention-weights are frozen during Prompt-to-Prompt inference. Through this work, we come to the interesting finding that instead of freezing the image-to-condition cross-attention weights, freezing image-to-image self-attention weights resulted in much more consistent geometry across synthetic pairs in pixel-level. Furthermore, inclusion of more scene conditions such as
agent 3D locations
and high-definition (HD) maps helps to improve the geometric consistency across paired synthetic images. 

Finally, we train a unified editing model: SceneCrafter, to jointly learn the aformentioned editing tasks. We propose novel 3D consistency metrics for multi-view image generation tasks, demonstrating significant realism, controllability, editing quality, and 3D consistency improvement on top of available baselines.

In summary, our key contributions are as follows:
\begin{itemize}
\item We present SceneCrafter, a multi-view driving scene editing model supporting local foreground object removal and injection, as well as global editing of operating conditions such as weather and time of day (See \figref{teaser}).
\item We propose a masked training paradigm and a multi-view repaint algorithm to remove agents from input images, along with an alpha-blending method for generating synthetic data, enabling training an editing model for flexible object insertion, removal, and replacement operations.
\item Our novel approach extends Prompt-to-Prompt~\cite{hertz2022prompt} for synthetic data generation in driving scenes. By integrating a modified attention-weight replacing mechanism and conditioning on scene layouts, we achieve significant improvements in both the realism and geometric consistency of generated data pairs across diverse operating conditions.
\item We propose \textit{3D LPIPS}, a metric to measure multi-view image consistency. Our results demonstrate a  marked increase in realism, controllability, 3D consistency, and editing quality over established baselines.
\end{itemize}

\section{Related Work}
\label{sec:prior}

\topic{Diffusion models.}
Diffusion models~\cite{sohl2015deep,song2019generative} have shown promising generation results in various domains, such as image~\cite{nichol2022glide,saharia2022photorealistic,betker2023improving,ramesh2022hierarchical}, video~\cite{ho2022imagen,gupta2023photorealistic,videoworldsimulators2024}, text~\cite{li2022diffusion}, and audio~\cite{kong2020diffwave}.
Latent diffusion models (LDMs)~\cite{rombach2022high} in particular mark a major milestone.
By learning the diffusion models in latent instead of pixel space, LDM greatly reduce the learning difficulty in the high-dimensional pixel space, and thus enable easier training and generation at higher resolutions.
In this work, we build our controllable multi-view scene editing model on top of the state-of-the-art LDM-based novel view synthesis model~\cite{gao2024cat3d}.

\topic{Image editing.}
Image editing is an important task in computer vision and graphics research.
In general, there are two ways to perform generative image editing: training-free and training-based.
Training-free approaches aim to leverage image priors from well-trained generative models without training a specific editing model.
To utilize the StyleGAN priors~\cite{karras2019style,karras2020analyzing}, previous approaches invert~\cite{abdal2019image2stylegan,abdal2020image2stylegan++,alaluf2022hyperstyle} or encode~\cite{chai2021using,richardson2021encoding,tov2021designing} the input images into the StyleGAN latent space and then perform editing by manipulating the latent vectors.
For diffusion models, Prompt-to-Prompt~\cite{hertz2022prompt} can edit generated images by updating the input text prompt and manipulating the cross attention weights so that the edit can be grounded to a specified region.
However, Prompt-to-Prompt mainly focuses on editing images generated by the text-conditioned diffusion model, instead of user-specified input images.
To edit real imagery, SDEdit~\cite{meng2021sdedit} uses a pre-trained diffusion model to add noise and then denoise the input images, but it struggles to maintain the fine-grained geometry of the edited results.
On the other hand, we can train specialist feed forward models to directly conduct the editing tasks.
In this work, we propose to generate training data with ideas from training-free approaches, and then train a unified editing model with synthetic data.

\topic{Synthetic training data generation.}
Training deep neural networks usually requires large amounts of data. However, collecting high-quality data at scale is not a trivial task and usually involves a significant amount of human efforts.
As generative models are evolving in visual fidelity, more and more research works~\cite{zhang2021datasetgan,li2023dreamteacher,wu2023diffumask,wu2023datasetdm} have been exploring training deep models with these generated synthetic data.
Our method is inspired by the pioneering work InstructPix2Pix~\cite{brooks2022instructpix2pix},
which leverages GPT3~\cite{mann2020language} and Prompt-to-Prompt to generate the editing prompts and training image pairs. We extend Prompt-to-Prompt for multi-view image editing, replacing text prompts with driving-specific signals for precise control. We also leverage Repaint algorithm to generate local editing data. Unlike DriveEditor~\cite{liang2024driveeditor}, which masks objects randomly for faster processing, our method removes identified vehicles, ensuring structured and realistic results.

\topic{Simulation for autonomous driving.}
Historically, most simulation works focus on driving policy simulation and scenario generation~\cite{caesar2021nuplan,montali2024waymo,gulino2024waymax}, which are then used to evaluate and improve performance of motion planners. However, this formulation fails to evaluate the whole AV system in a closed-loop setting.
Given the recent trend of end-to-end driving models~\cite{hu2022st,hu2023planning,gu2023vip3d,hwang2024emmaendtoendmultimodalmodel}, how to realistically generate sensor data for full system closed-loop simulation becomes a critical problem.
Recently generative world models~\cite{hu2023gaia1generativeworldmodel, ma2024unleashinggeneralizationendtoendautonomous, zhao2024drive, wang2023drivedreamerrealworlddrivenworldmodels, wang2024driving, gao2024vista, li2023drivingdiffusion, yang2024genad, gao2023magicdrive, yang2023unisim, wei2024editable} are capable of generating photo-realistic future frames, conditioned on language or action prompts. Panacea~\cite{wen2024panacea} can even generate multi-view videos given vehicle actions, agent locations, and the environment map.
However, these methods mainly focus on generating the future or a completely new scene, but not editing real driving footage.
In this paper, we aim to fill this gap by proposing a versatile editor for realistic 3D consistent driving scene editing.

\vspace{-2mm}

\section{Method}
\label{sec:method}

In this section, we first provide a preliminary review of the multi-view diffusion model in \secref{pre}.
We then provide details on the architecture in \secref{teacher} and explain the process of creating synthetic datasets using the teacher model in \secref{generate_data}. Specifically, we train two separate teachers models to generate data for global edits and local edits, respectively. Lastly, in \secref{edit} we describe how to train a unified student editing model with the generated paired data. An overview of our pipeline for synthetic dataset generation and the controllable diffusion framework is illustrated in \figref{method}.

\subsection{Preliminary: Multi-View Diffusion Models}
\label{sec:pre}
Our method is based on multi-view diffusion models~\cite{gao2024cat3d}, which extends the latent diffusion models (LDM)~\cite{rombach2022high} by enabling view consistency and camera pose control over the image generation. The goal of this architecture is to estimate the joint distribution of multi-view images under given camera poses, and generate photo-realistic and view-consistent images. Specifically, given a set of $N$ camera poses $\mathbf{p} = p_{0:N}$ where $N = 8$ in our setting, the model learns to estimate the distribution of $N$ images $\mathcal{I} =I_{0:N}$ under the corresponding camera poses:
$P(\mathcal{I}|\mathbf{p})$.

Following \cite{rombach2022high}, the base architecture is a latent diffusion model with an encoder $\mathcal{E}$, a denoiser U-Net $\epsilon_{\theta}$ and a decoder $\mathcal{D}$. The images are encoded into latents $\mathbf{z} = z_{0:N}$ and then denoised in latent space:
$\mathcal{I} = \mathcal{D}(\mathbf{z}), \mathbf{z} = \mathcal{E}(\mathcal{I})$.

Two key architecture changes are adopted to achieve the aforementioned goal: a view-spatial joint attention module and camera pose conditioning generation.

\topic{View-spatial joint attention}. The 2D attention blocks in the LDM are replaced with 3D attention blocks (2D in space and 1D cross views) to perform the attention mechanism among multi-view images. To reuse the parameters from the original LDM, the 2D attention module is directly inflated to 3D without introducing extra parameters.
All other 2D blocks are applied to each image separately.

\topic{Camera poses conditioning}.
The camera poses are represented via raymaps, which encode the ray origin and direction at each spatial location. All camera poses are normalized w.r.t. the first camera so the raymaps are invariant to global rigid transformations. The diffusion U-Net takes the the concatenated noisy latents and raymaps as input, and outputs the denoised latents.

\subsection{Teacher Model for Scene Generation}
\label{sec:teacher}
In this section, we introduce how we accommodate various conditional modalities in the teacher model. 
The model estimates the multi-view image distribution given groups of conditions as $
    p(\mathcal{I}|\mathbf{p}, \mathbf{m}, \mathbf{c}^g, \mathbf{c}^l)
$.

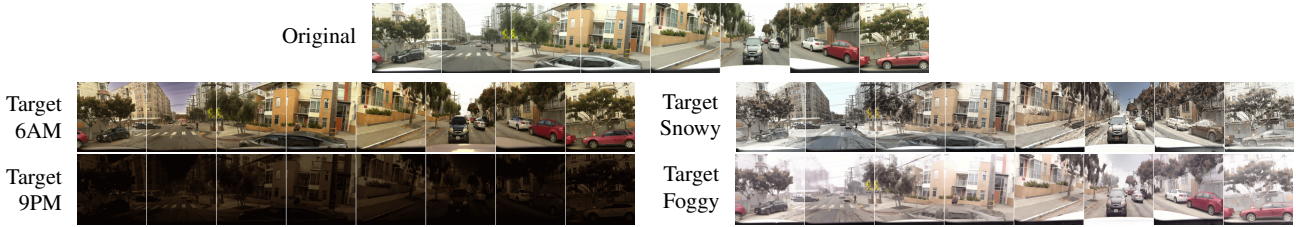
\begin{figure*}[t]
    \centering

\resizebox{\linewidth}{!}{
\begin{tikzpicture}[node distance=0cm]

\def\imagewidth{8.8cm}
\def\imageheight{1.1cm}
\def\verticalgap{0.5cm} 

\node (start) {};
\node[below=0.9cm of start] (start2) {};

\node[above=0.45cm of start, anchor=south] (image1) {\includegraphics[width=\imagewidth, height=\imageheight]{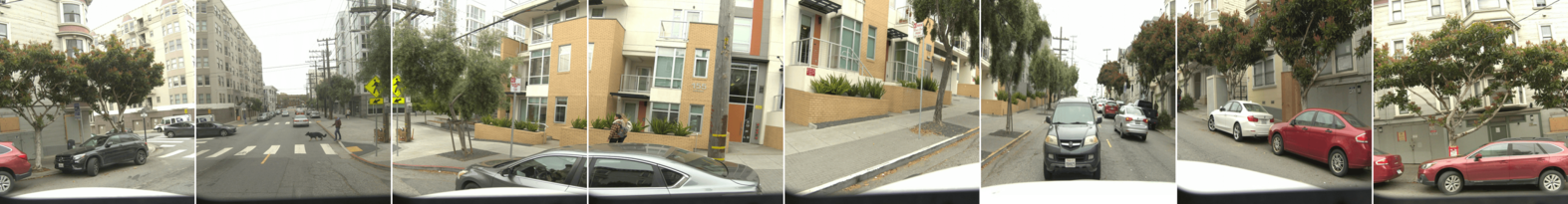}}; 
\node[anchor=east] at (image1.west) {Original}; 

\node[left=of start, anchor=east] (image2) {\includegraphics[width=\imagewidth, height=\imageheight]{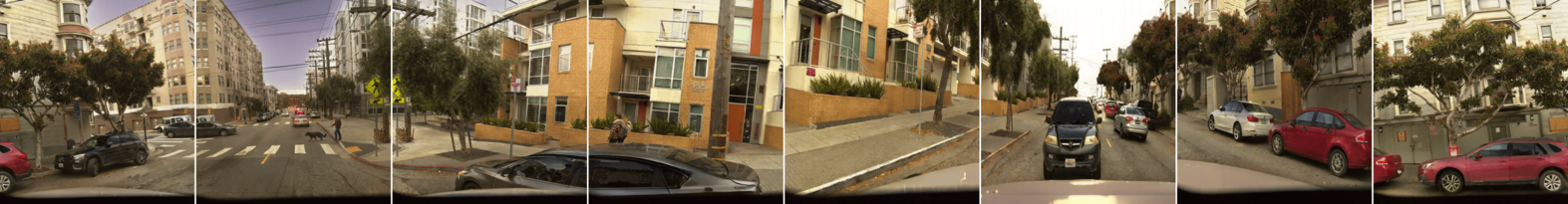}}; 
\node[left=of image2, anchor=east, align=right] (a) {Target\\6AM}; 
\node[right=-0.1cm of start, anchor=west, align=center] (b) {Target\\Snowy}; 
\node[right=of b, anchor=west] (image3) {\includegraphics[width=\imagewidth, height=\imageheight]{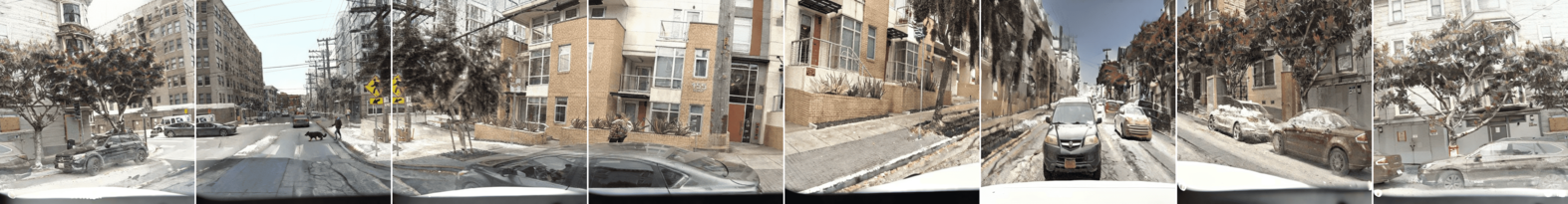}}; 

\node[left=of start2, anchor=east] (image4) {\includegraphics[width=\imagewidth, height=\imageheight]{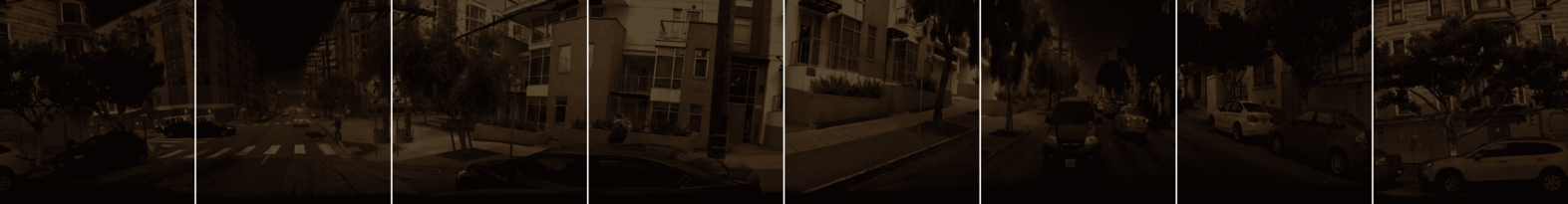}}; 
\node[left=of image4, anchor=east, align=right] (d) {Target\\9PM}; 
\node[right=-0.05cm of start2, anchor=west, align=center] (e) {Target\\Foggy}; 
\node[right=of e, anchor=west] (image5) {\includegraphics[width=\imagewidth, height=\imageheight]{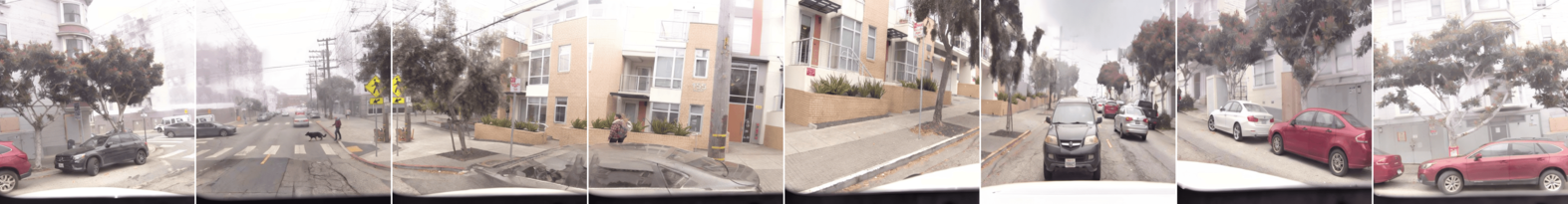}}; 

\end{tikzpicture}
}
\vspace{-8mm}
    \caption{\textbf{Qualitative Results on Global Editing.}
    Given multi-view image inputs, our model performs versatile edits like changing the time of day (daytime to dawn/night) and weather (sunny to snowy/foggy) while preserving geometric consistency. \textit{Best viewed zoomed in.}
    }
    \label{fig:edit}
\end{figure*}
Specifically, the diffusion model takes four types of conditions: global conditions $\mathbf{c}^g$ (weather and time of day), local conditions $\mathbf{c}^l$ (HD map and agent boxes), foreground masks $\mathbf{m}$ and raymaps $\mathbf{p}$. The global conditions are essential for simulation under novel operating conditions, and the latter two help the model capture the scene layout and geometry. We encode all the conditions and integrate them into the cross-attention blocks of the diffusion U-Net.

\noindent \textbf{Weather.}
Based on the driving log data, we use the CLIP~\cite{radford2021learning} text encoder to encode the text ``sunny", ``rainy", ``foggy", or ``snowy" as the weather condition $c_w$.

\noindent \textbf{Time of day.} 
Given the local time of day and the geographic location of the recorded driving logs, we compute the sun angles for each frame, which is then encoded using positional encoding and used as time of day condition $c_t$.

\noindent \textbf{High-definition map.}
We also utilize a high-definition (HD) environment map and process it to the local HD map conditions $c_r$.
The map is represented by lane segments and lane types. We sample up to 4,096 lane segments, feed the segment start and end locations, and lane types to a PerceiverIO~\cite{jaegle2022perceiver} to reduce the token size to 512, and use a MLP to encode the features into the condition $c_r$. 

\noindent\textbf{Agent boxes.}
We use the AV perception system's 3D object detector to gather up to 256 agents in the scene, including both foreground and background objects. Each agent is represented by an 8-dimensional feature tensor comprising the center coordinates (\(x, y, z\)), dimensions (length, width, height), heading angle (yaw), and agent type.
We apply an one-hot encoding of agent type and concatenate it with the remaining features. We then feed the concatenated features to a MLP to get output $c_b$ as the agent condition.

\noindent\textbf{Foreground mask.}
We project all 3D bounding boxes of foreground objects into each camera view to generate binary masks. We then resize them and append them to the latents along the channel dimension similar to raymaps.

We apply a 10\% dropout rate to each conditions during training. So our model does not rely on these conditions and works robustly without them.

\subsection{Synthetic Data Generation with Teacher Model}
\label{sec:generate_data}

\subsubsection{Generating Data for Global Edits}
\label{sec:global_edits}

To generate paired data for global edits (time of day and weather), an intuitive idea is to feed the initial and edited conditions into a well-trained conditional generation model to generate the \textbf{source} and \textbf{target} images respectively.
However, this simple strategy fails to generate geometrically consistent source and target image pairs, as shown in the first and second rows of \figref{ablation_p2p}.
This is because generative models inherently lack guarantees for image coherence, even when only minor changes are made to the conditioning prompt~\cite{brooks2022instructpix2pix}.
To address this issue, we employ Prompt-to-Prompt~\cite{hertz2022prompt}, a method designed to maintain similarity across multiple outputs of a conditional diffusion model. 
However, the original Prompt-to-Prompt approach falls short in terms of controllability and realism for our multi-view driving scene setting.
Thus, we make the following adaptation to Prompt-to-Prompt.

\topic{Replacing self-attention weights.} Different from the original Prompt-to-Prompt approach that manipulates image-to-condition cross-attention weights, we extend this to image-to-image self-attention layers to better handle global conditions.
While Prompt-to-Prompt uses text tokens and substitutes semantically meaningful text tokens, our setting handles various multi-modal tokens and replaces certain tokens associated with global effects, while keeping other conditions unaffected. The editing should affect all parts of the image, but
still retain the original layout, such as the location and underlying appearance of the vehicles. Therefore, we manipulate pixel-level attention weights in all self-attention layers instead of the cross-attention layers.

\topic{Conditioning on more control signals.}
We observed that incorporating more control signals into our model enables generating more realistic paired results with consistent geometry. With more control signals, the model better preserves fine-grained details which ground the geometry specifically while primarily modifying high-level attributes such as weather. So we incorporate more local conditions into our teacher model, like agent boxes and HD maps.

\topic{Choice of source images.} The time of day choice for source images plays a crucial role in our method, as it impacts the overall style of the generated results. We empirically found that using only daytime for source images produces superior results compared to using nighttime or using both. Thus we sample the time of day for source images from daytime only and sample target time from all times across a day. We randomly flip the order of source and target images as the paired synthetic data.

\subsubsection{Generating Data for Local Edits}
\label{sec:local_edits}
In local edits, we aim to remove or insert arbitrary agents into the scene. Our log data is populated with vehicles, but if we can extract ``empty streets" data from it, we can then perform alpha blending for the pairs, which enables generating synthetic data with arbitrary numbers of agents. 
Inpainting methods like RePaint~\cite{lugmayr2022repaint} have been widely applied to erase objects in 2D images, however it faces two challenges in our settings: It can not produce multi-view consistent results;
and our ``populated streets" training data makes it difficult to ensure inpainted results are free of agents when directly training a diffusion model.
Thus, we propose a novel framework with two key ideas: \textit{masked training} and \textit{multi-view repaint} to tackle these issues. Masked training enables the model to learn priors of empty streets in a self-supervised manner and multi-view repaint ensures view-consistent results. 

\topic{Masked training.}
We propose a simple yet effective training strategy that enables our model to learn the ``empty street" prior.
Specifically, given a set of image foreground mask for each camera views, we resize them to the same shape with latents, denoted as $\mathbf{m}=m_{0:N}$. 
We denote foreground latents as $\mathbf{m} \odot \mathbf{z}$ and background latents as $(1 - \mathbf{m}) \odot \mathbf{z}$. 

We apply different noise levels when computing $\mathbf{z}_t$. We maintain a noise level of zero for the foreground and apply regular noise to background, as:
\vspace{-2mm}
\begin{equation}
\mathbf{z}_t = (1 - \mathbf{m}) \odot (\alpha_t \mathbf{z}_0 + \sigma_t \epsilon) + \mathbf{m} \odot \mathbf{z}_0
\vspace{-2mm}
\end{equation}
where $\alpha_t$ and $\sigma_t$ are predefined noise scheduling terms. Consequently the model focuses on learning to denoise the background while leaving the foreground unchanged.

We only compute training loss on background pixels, and minimize the following training objective,
\vspace{-2mm}
\begin{equation}
\hspace{-1mm}
\small
\mathbb{E}_{z \sim \mathcal{E}(I), t, \epsilon \sim \mathcal{N}(0, 1)} \left\| (1 - \mathbf{m}) \odot (\epsilon - \epsilon_{\theta}(\mathbf{z}_t, t, \mathbf{c}^g, \mathbf{c}^l, \mathbf{m})) \right\|_2^2
\vspace{-2mm}
\end{equation}

The model $\epsilon_{\theta}$ is trained with foreground-free priors and thus can generate scenes of ``empty streets".

\begin{table*}[t]
\large
\centering
\vspace{-1mm}
\resizebox{1.5\columnwidth}{!}{
\begin{tabular}{l|ccc|ccc} 
\toprule
\multirow{2}{*}{{Methods}} & \multicolumn{3}{c|}{Time of day editing} & \multicolumn{3}{c}{Weather editing} \\ 
                                  & FID$\downarrow$ & CLIP Score$\uparrow$ & User Study$\uparrow$ & FID$\downarrow$ & CLIP Score$\uparrow$ & User Study$\uparrow$ \\ 
\midrule
SDEdit~\cite{meng2021sdedit}                       &    60.4             &       0.204               &      2.7\%                  &    78.3              &         0.203             &     1.8\%                \\
P2P*~\cite{hertz2022prompt}       &  46.8                  &    \textbf{0.223}            &        13.6\%              &       55.4               &          0.207                           &         12.7\%              \\
SceneCrafter                              &    \textbf{37.2}              &     0.220                &       \textbf{83.6\%}                  &     \textbf{38.9}            &     \textbf{0.221}                 &            \textbf{85.5\%}           \\ 
\bottomrule
\end{tabular}
}
\vspace{-3mm}
\caption{
\textbf{Quantitative Comparison on Global Editing.}
SceneCrafter shows clear improvement over two baselines in terms of realism, controllability and editing quality, achieving preference rates of over 80\% across both editing benchmarks. 
}
\vspace{-2mm}
\label{tab:edit}
\end{table*}
\topic{Multi-view repaint.}
We propose a 3D-aware approach to get view-consistent inpainting examples, replacing foreground pixels with backgrounds.
Note that we only conduct multi-view repaint with the teacher model with masked training.
Specifically, we split the images into masked and unmasked regions representing foreground and background, respectively.
In each reverse step, we modify the foreground region \(\mathbf{m} \odot \mathbf{z}_t\) while preserving the correct properties of the corresponding distribution. Background region \((1-\mathbf{m}) \odot \mathbf{z}_t\) is sampled at any time step \(t\) given known latents \(\mathbf{z}_0\). Thus, we have:
\vspace{-1mm}
\begin{align}
  \mathbf{z}_{t-1}^{\text{background}} &\sim \mathcal{N}\left(\sqrt{\bar{\alpha}_t} \mathbf{z}_0, (1 - \bar{\alpha}_t) \mathbf{I}\right) \\
  \mathbf{z}_{t-1}^{\text{foreground}} &\sim \mathcal{N}\left(\mu_{\theta}(\mathbf{z}_t, t, \mathbf{c}^g, \mathbf{c}^l), \Sigma_{\theta}(\mathbf{z}_t, t, \mathbf{c}^g, \mathbf{c}^l)\right)
\end{align}

Here, \(\mathbf{z}_{t-1}^{\text{background}}\) is sampled based on the known pixels in the given image, while \(\mathbf{z}_{t-1}^{\text{foreground}}\) for all camera views is sampled simultaneously from the multi-view diffusion model, using the previous iteration \(\mathbf{z}_t\). 
Finally, we merge these two latents where the foreground regions are progressively denoised to background:
\vspace{-2mm}
\begin{equation}
\mathbf{z}_{t-1} = \mathbf{m} \odot \mathbf{z}_{t-1}^{\text{foreground}} + (1 - \mathbf{m}) \odot \mathbf{z}_{t-1}^{\text{background}}
\end{equation}
\vspace{-4mm}

\topic{Alpha blending.}
After applying multi-view repaint to ``populated streets" data, we obtain the corresponding ``empty streets" data, denoted as \(\mathcal{I}^{\text{empty}}\) and \(\mathcal{I}^{\text{full}}\).
We then employ alpha blending to generate paired data for object insertion or removal.
More specifically, 
we first project all agent bounding boxes onto 2D planes to create a vehicle mask.
We then sample any desired number of these masks to form a new composite mask, \(\mathbf{m}^{\text{sampled}}\), which contains varying numbers of agents.

The alpha blending of \(I^{\text{empty}}\) and \(\mathcal{I}^{\text{full}}\) is performed as:
\vspace{-1mm}
\begin{equation}
  \mathcal{I}^{\text{sampled}} = \mathbf{m}^{\text{sampled}} \odot \mathcal{I}^{\text{empty}} + (1 - \mathbf{m}^{\text{sampled}}) \odot \mathcal{I}^{\text{full}}
\vspace{-1mm}
\end{equation}
which allows us to seamlessly blend the images, creating realistic scenes with any specified number of agents.

\subsection{Student Model for Scene Editing}
\label{sec:edit}

We frame the image-to-image editing as a generation task, conditioning on source images $\mathcal{I}^{\text{source}}$ and the control signals. To effectively incorporate the source images into the model, we introduce an additional conditioning branch that concatenates the latent of the source images $\mathbf{z}_0^{\text{source}}$  directly with the  latents $\mathbf{z}_t$. We find that concatenating pixel-level features, such as source images, masks, and raymaps yields better results than using cross-attention techniques, as shown in Sec.~\ref{sec:ablation}.

For training the editing model, we employ the synthetic dataset detailed in Sec.~\ref{sec:generate_data}, which includes changes in weather, time of day and agents boxes. The editing model's weights are initialized from the weights of the global editing teacher model. We maintain the original conditioning mechanisms of the generation model and also update their parameters during training. This integration ensures that the editing model can leverage the detailed contextual data from the source images while maintaining the generative capabilities of the original model.

At test time, our model takes source images and arbitrary target prompts, then generates target images that preserve the geometric structure and layout of the source images while aligning with the specified prompts. It's worth noting that our model uses box conditions for agent editing rather than masks, and agent insertion or removal are controlled by agent types. 
Unlike mask-based methods that often produce imprecise boundaries, our box conditioning offers more precise control, especially for smaller objects illustrated in \figref{teaser}. This allows us to achieve fine-grained edits while maintaining geometric consistency with the original image.

\section{Experiments}
\label{sec:exp}

\begin{figure*}[t]
    \centering
    \begin{minipage}[t]{0.49\linewidth} 
        \centering
        \resizebox{\linewidth}{!}{ 
        \begin{tikzpicture}
        \def\imagewidth{15cm} 
        \def\imageheight{2.2cm} 
        \def\verticalgap{-0.2cm} 

        \node (image1) {\includegraphics[width=\imagewidth, height=\imageheight]{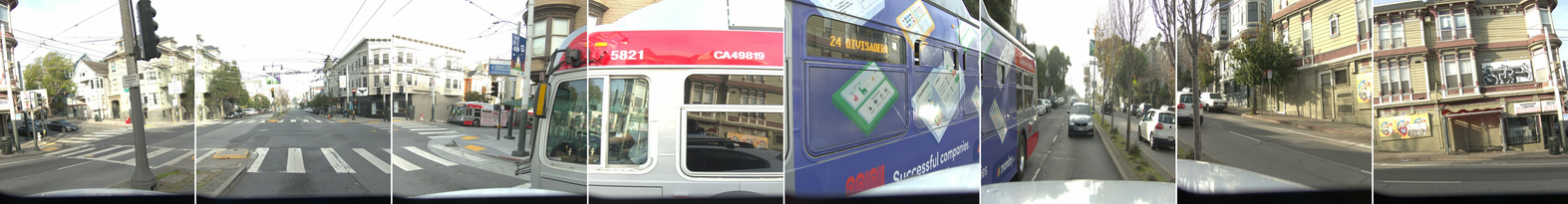}}; 
        \node[anchor=east] at (image1.west) {Source}; 

        \node[below=\verticalgap of image1] (image2) {\includegraphics[width=\imagewidth, height=\imageheight]{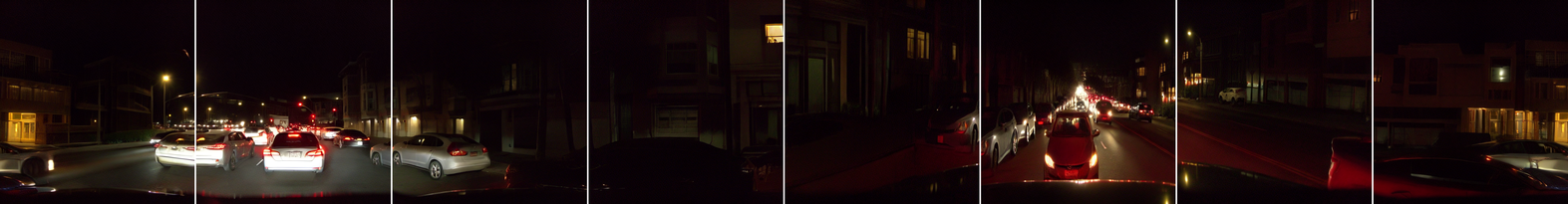}}; 
        \node[anchor=east, align=right] at (image2.west) {SDEdit~\cite{meng2021sdedit}}; 

        \node[below=\verticalgap of image2] (image3) {\includegraphics[width=\imagewidth, height=\imageheight]{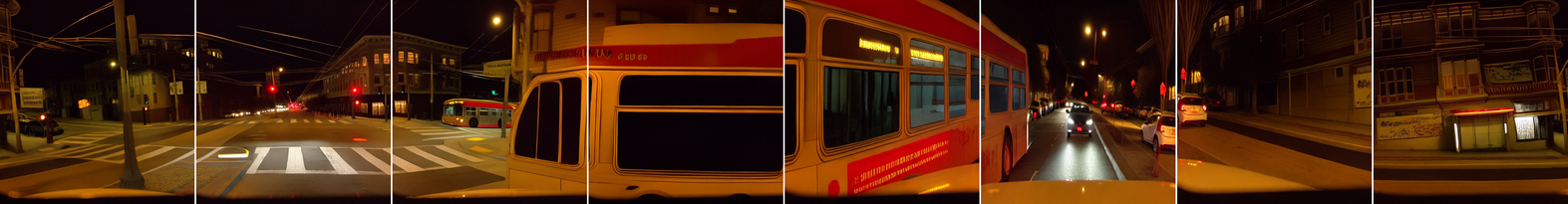}}; 
        \node[anchor=east, align=right] at (image3.west) {P2P*~\cite{hertz2022prompt}}; 

        \node[below=\verticalgap of image3] (image4) {\includegraphics[width=\imagewidth, height=\imageheight]{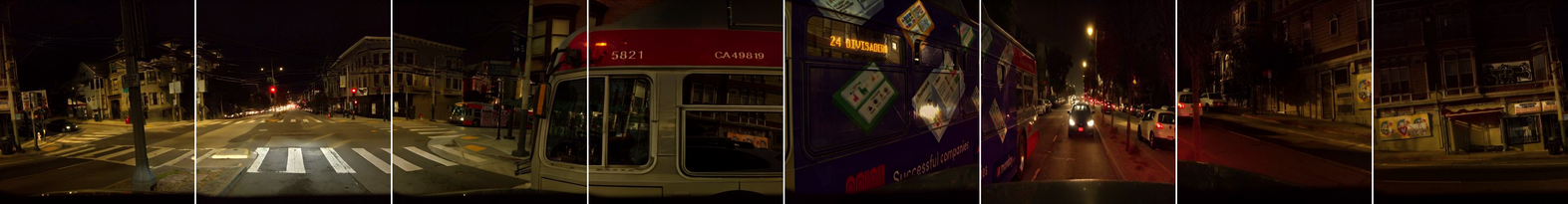}}; 
        \node[anchor=east, align=right] at (image4.west) {SceneCrafter};
        \end{tikzpicture}
        }
        \vspace{-6mm}
        \caption{
        \textbf{Qualitative Comparison on Time of Day Editing.}
        SceneCrafter is able to conduct realistic editing to 8PM while still maintaining the geometric structure of the input images.
        }
        \label{fig:compare_time}

    \end{minipage}%
    \hfill
    \begin{minipage}[t]{0.49\linewidth} 
        \centering
        \resizebox{\linewidth}{!}{ 
        \begin{tikzpicture}
        \def\imagewidth{15cm} 
        \def\imageheight{2.2cm} 
        \def\verticalgap{-0.2cm} 

        \node (image1) {\includegraphics[width=\imagewidth, height=\imageheight]{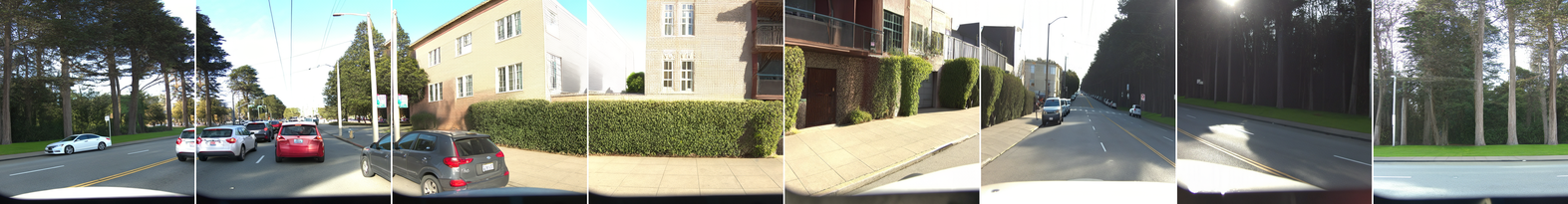}}; 
        \node[anchor=east] at (image1.west) {Source: 8AM}; 

        \node[below=\verticalgap of image1] (image2) {\includegraphics[width=\imagewidth, height=\imageheight]{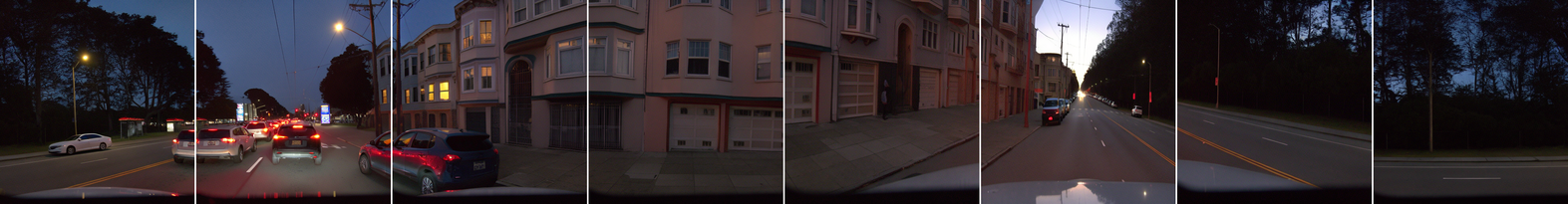}}; 
        \node[anchor=east, align=right] at (image2.west) {Target: 7PM\\(w/o P2P.)}; 

        \node[below=\verticalgap of image2] (image3) {\includegraphics[width=\imagewidth, height=\imageheight]{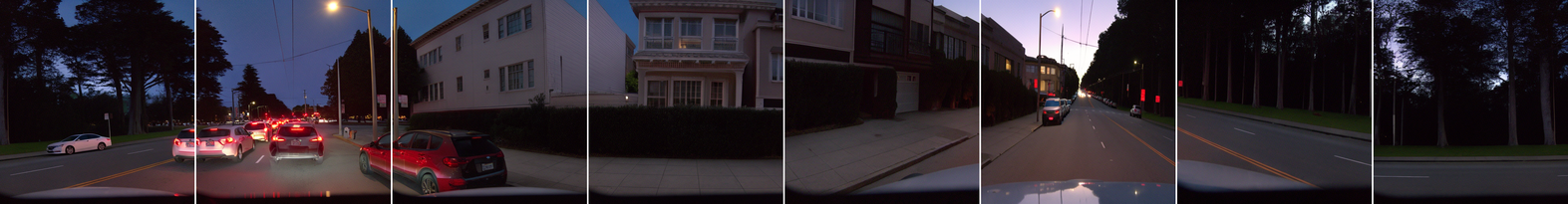}}; 
        \node[anchor=east, align=right] at (image3.west) {Target: 7PM\\(w/ cross.)}; 

        \node[below=\verticalgap of image3] (image4) {\includegraphics[width=\imagewidth, height=\imageheight]{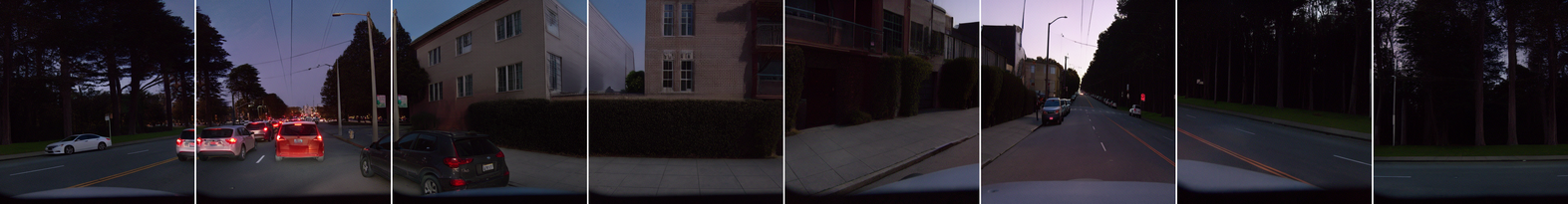}}; 
        \node[anchor=east, align=right] at (image4.west) {Target: 7PM\\(w/ self.)};
        \end{tikzpicture}
        }
        \vspace{-6mm}
        \caption{
        \textbf{Attention Weights Manipulation in Prompt-to-Prompt.}
        Manipulating self-attention weights enables accurate conditioning while preserving the same geometry.
        }
        \label{fig:ablation_p2p}
    \end{minipage}
\end{figure*}

\subsection{Metrics}

Evaluating generative 3D editing models presents unique challenges. Metrics for novel view synthesis typically emphasize generation quality, but additional criteria are essential for 3D editing tasks. Specifically, we assess models based on realism, controllability, 3D consistency and editing quality. We utilize existing metrics such as Fréchet Inception Distance (FID)~\cite{DBLP:conf/nips/HeuselRUNH17} and CLIP Score~\cite{DBLP:conf/emnlp/HesselHFBC21}, run a user study, and propose a novel 3D LPIPS metric to comprehensively assess these aspects.

\topic{Realism.} FID is the most commonly used metric for estimating the realism of generated images by measuring how similar the distribution of edited images is to the distribution of the origin images in a feature space.

\topic{Controllability.} The CLIP score originally aims to measure how well an image caption aligns with the semantics of an image.
We adapt CLIP Score to evaluate controllability by converting a control signal into a text to describe how to edit a image, then calculate the cosine similarity between the text and edited image embeddings.

\topic{Editing quality.} We conduct a user study to evaluate the editing quality of our method compared to baselines. The user study involved 11 human raters evaluating 20 groups of images, where each group corresponds to a unique edit prompt. For each group, we show users the images generated by three methods (randomized order) and the source images. Users were asked \textit{``which of these images is the most faithful result of editing the source image according to the given prompt?"} and then selected the image they thought most aligned with the question. We compiled their selections to summarize user preferences.

\begin{table}[t]
\vspace{-3mm}
\centering
\resizebox{0.70\linewidth}{!}{
\begin{tabular}{l|cc} 
\toprule
\multirow{2}{*}{{Method}} & \multicolumn{2}{c}{FID$\downarrow$} \\ 
\cline{2-3}
                                  & Removal & Insertion  \\ 
\hline
2D-RePaint~\cite{lugmayr2022repaint}                       &        30.6         &     31.9     \\
MV-RePaint                           &       26.0          &      28.5        \\
SceneCrafter                            &        \textbf{23.5}          &     \textbf{21.7}       \\ 
\bottomrule
\end{tabular}}
\vspace{-2mm}
\caption{
\textbf{Quantitative Comparison on Local Editing.} SceneCrafter, which uses agent boxes to manipulate agents, shows obvious advantages over two mask-based methods, which often struggle with imprecise segmentation, where masks exceed object boundaries. Our box-based conditioning enables more accurate, fine-grained edits, especially for small objects.
}
\label{tab:inpaint}
\end{table}

\topic{3D Consistency.} We leverage the overlapping field of view between cameras to measure consistency across multiple views. More specifically, for each neighboring view pair $(C_i, C_{i+1})$, we project $C_i$ into $C_{i+1}$, as well as $C_{i+1}$ into $C_i$, and compare the overlapping regions. We use the LPIPS metric as it is known to be sufficiently robust to effects such as exposure differences and motion blur~\cite{zhang2018unreasonable}, and subsequently also refer to this metric as 3D LPIPS.

\subsection{Experimental Settings}
\topic{Dataset.}
Since our multi-view driving scene editing and generation tasks are unique, we require a novel dataset to enable control over both global and local scene conditions, as well as camera poses. 
We curated a dataset that consists of 13,867,496 unique segments of driving videos for training the teacher diffusion models. Each segment consists of 17 frames of 8 surrounding cameras, captured in the frequency of 10 Hz. The log data contains versatile labels such as camera poses, weather, time of day, HD map, and agent bounding boxes, estimated by the AV onboard stack. These labels enable our teacher model to associate control signals with scene generation.
We held out 1\% videos for testing and used the remaining data for teacher model training.

\topic{Implementation details.} We base our two teacher models for global and local edits on a pre-trained multi-view diffusion model~\cite{gao2024cat3d} and fine-tune the global teacher model for final student editing model. We train our models on 128 Google TPU v5 for 100k iterations, with learning rate of $1e^{-5}$ and batch size 128. We generate 1M synthetic paired data to train the student model.
We resize all the inputs to $512 \times 512$ to align with the pre-trained VAE.
At inference time, we use 50 denoising steps with classifier-free guidance~\cite{ho2020denoising}.

\begin{table}[t]
\centering
\vspace{-3mm}

\resizebox{0.8\linewidth}{!}{
\begin{tabular}{l|ccc} 
\toprule
{Method} & FID $\downarrow$ & 3D LPIPS $\downarrow$ \\
\midrule
Real & 11.5 & 0.186\\ 
CAT3D~\cite{gao2024cat3d} & 121.3 & 0.249\\ 
SceneCrafter (w/o cond.)& 68.5 & 0.254\\ 
SceneCrafter (full) & \textbf{36.2} & \textbf{0.187} \\ 
\bottomrule
\end{tabular}}

\vspace{-2mm}
\caption{
\textbf{Quantitative Results on Generation Task.}
With conditioning signals encoded, SceneCrafter can generate more realistic multi-view images compared to the baselines.
SceneCrafter can even achieve the same degree of 3D consistency as real log data, measured by the novel 3D LPIPS metric.
}
\label{tab:gen}
\vspace{-3mm}
\end{table}

\subsection{Results}

\topic{Qualitative editing results.} We demonstrate our qualitative editing results in \figref{edit}.
Given arbitrary multi-view source images as input (first row), our model can perform many challenging edits, including changing time from day to dawn and night (left), and changing weather from sunny to snowy and foggy (right), while keeping the scene geometrically consistent.
We show more creative editing results in the supplemental material.

\topic{Comparison to other editing baselines.} For global edits, we provide quantitative comparisons with SDEdit~\cite{meng2021sdedit} and P2P* in \tabref{edit}, along with qualitative comparisons on time of day editing in \figref{compare_time}. 
SDEdit is a general image editing method where a partially noised image is denoised to generate a new edited one. We extend Prompt-to-Prompt for editing tasks by replacing the latents of the source images with that of the images to be edited, and the generated target images are the image after editing, denoted as P2P*.
In \tabref{edit}, we observe that our realism, controllability and editing quality are consistently better than other baselines in all but one instance. SceneCrafter outperforms SDEdit and P2P$^{*}$ by 40.6 and 16.5 in FID and achieves a 83.6\% and 85.5\% preference rating from our user study, respectively, demonstrating that our method conditions precisely on the source images and achieves accurate editing. Visually, as shown in \figref{compare_time}, our method excels at generating consistent geometry while keeping highly detailed textures. Other baselines provide results aligned to the target prompts, but exhibit sub-optimal geometric consistency.

We also compare our method against 2D-RePaint and MV-RePaint for local editing tasks, as shown in \tabref{inpaint}. For the 2D-RePaint baseline, we utilize a pretrained Stable Diffusion model~\cite{rombach2022high}, applying it independently to each camera view. For the MV-RePaint baseline, we use our editing model without conditioning on source images and adopt the multi-view repaint method in \secref{local_edits}, using 2D-projected agent bounding boxes as masks.
We evaluated on agent insertion and removal scenarios. Our method consistently demonstrated superior object editing quality compared to both approaches, and achieves the best FID scores. 2D RePaint lacks editing type control and MV-RePaint, on the other hand, relies heavily on masks, which might not always accurately align with object boundaries. Our method conditions on specific agent bounding boxes, allowing precise insertion and removal.

\begin{table}[t]
\centering
\resizebox{\linewidth}{!}{
\begin{tabular}{ccccc}
\toprule
Cross   & Increase  &  Daytime for  & \multirow{2}{*}{FID$\downarrow$}  & \multirow{2}{*}{CLIP Score$\uparrow$}  \\
Attention  & Conditioning & Source Images  &       \\  
\midrule
\xmark & \xmark & \xmark &   57.1   &  0.204   \\
\cmark & \xmark & \xmark  &  41.5   &  0.202  \\
\cmark & \cmark & \xmark  &  39.9    &  0.214   \\
\cmark & \cmark & \cmark  &   \textbf{36.2}  &  \textbf{0.223}   \\
\bottomrule
\end{tabular}
}
\vspace{-0.7em}
\caption{{\bf Ablation Study on Prompt-to-Prompt Design.} We show three key components to improve the quality of synthetic data. First, replacing self-attention weights led to better geometric consistency for generated pairs. Second, using additional conditions (agent features and HD maps) improved controllability. Finally, using only daytime source images enhanced the generation quality by a large margin.
}\label{tab:ablation_p2p}
\vspace{-3mm}
\end{table}

\topic{Comparison to other generation baselines.} To validate that our editing model also possesses generation ability, we present some pure generation results in \tabref{gen}. We compare with CAT3D~\cite{gao2024cat3d} by using a single view as the conditional input and generating the other seven views.
We also test our SceneCrafter without any conditions.
Notably, the full SceneCrafter model surpasses the baselines across both metrics. Moreover, our quantitative results closely match real image data, particularly in the 3D LPIPS metric. This superior performance can be attributed to the use of local conditions, which effectively grounds the geometric structure of 3D scenes and enables the creation of highly 3D consistent scenes well aligned with the real-world. In contrast, the other two baselines lack such geometrical constraints and show lower 3D consistency.

\subsection{Ablation Study}
\label{sec:ablation}
We ablate our design choices for Prompt-to-Prompt in \tabref{ablation_p2p} and different ways for source image condition in Tab.~\ref{tab:ablation_condition}.
We conduct ablation studies with time of day edits.

\topic{Design choices for prompt-to-prompt.} \tabref{ablation_p2p} demonstrates the improvement on generating synthetic data by introducing several key components, such as replacing self-attention weights, including more conditions and using only daytime for source images. Starting from the original Prompt-to-Prompt (first row), we find that replacing self-attention weights results in more consistent geometry in pixel-level image-to-image translation (second row). Including more scene conditions such as agent boxes and HD maps also helps to improve the condition quality across paired synthetic images (third row). Using daytime as the time of day for source images further enhances the generation quality and controllability (fourth row).

We also show qualitative visualizations of the effects of replacing different attention layers in ~\figref{ablation_p2p}. We compare not replacing any attention layers, replacing cross-attention layers, and replacing self-attention layers. We find that replacing self-attention layers only preserves the geometric consistency of the generated pairs.

\topic{Source image conditioning.}
We explore methods for conditioning source images in \tabref{ablation_condition}. For cross-attention, we encode images using the VAE and feed them directly, along with other conditions, into the cross-attention module. We find that concatenation yields better performance, notably improving FID by 13.1. This suggests that concatenation is more effective for pixel-level features like source images, masks, and raymaps, while cross-attention performs better with global and local conditions.

\begin{table}[t]
\centering
\resizebox{0.7\columnwidth}{!}{
\begin{tabular}{l|ccc} 
\toprule
{Method} & FID $\downarrow$ & CLIP Score $\uparrow$\\
\midrule
Cross-attention & 50.3 & 0.203 \\
Concatenation & \textbf{37.2} & \textbf{0.220} \\
\bottomrule
\end{tabular}}
\vspace{-2mm}
\caption{
\textbf{Ablation Study on Source Image Conditioning.}
Concatenating the multi-view source images clearly outperforms applying the cross-attention operation as woth other conditions.
}
\label{tab:ablation_condition}
\end{table}

\section{Conclusion}
\label{sec:conclusion}
We present SceneCrafter, a versatile editor for realistic 3D consistent multi-view driving scene image editing.
We decompose the problem into two steps.
First, we train teacher models to generate high-quality synthetic data using novel training paradigms. Next, we distill knowledge from these teacher models by training a unified student model on the generated dataset. We show that our student model is able to conduct several challenging editing tasks, with either global or local editing prompts.

\clearpage
{
    \small
    \bibliographystyle{ieeenat_fullname}
    \bibliography{main}

\begin{thebibliography}{59}
\providecommand{\natexlab}[1]{#1}
\providecommand{\url}[1]{\texttt{#1}}
\expandafter\ifx\csname urlstyle\endcsname\relax
  \providecommand{\doi}[1]{doi: #1}\else
  \providecommand{\doi}{doi: \begingroup \urlstyle{rm}\Url}\fi

\bibitem[Abdal et~al.(2019)Abdal, Qin, and Wonka]{abdal2019image2stylegan}
Rameen Abdal, Yipeng Qin, and Peter Wonka.
\newblock Image2stylegan: How to embed images into the stylegan latent space?
\newblock In \emph{ICCV}, 2019.

\bibitem[Abdal et~al.(2020)Abdal, Qin, and Wonka]{abdal2020image2stylegan++}
Rameen Abdal, Yipeng Qin, and Peter Wonka.
\newblock Image2stylegan++: How to edit the embedded images?
\newblock In \emph{CVPR}, 2020.

\bibitem[Alaluf et~al.(2022)Alaluf, Tov, Mokady, Gal, and Bermano]{alaluf2022hyperstyle}
Yuval Alaluf, Omer Tov, Ron Mokady, Rinon Gal, and Amit Bermano.
\newblock Hyperstyle: Stylegan inversion with hypernetworks for real image editing.
\newblock In \emph{CVPR}, 2022.

\bibitem[Betker et~al.(2023)Betker, Goh, Jing, Brooks, Wang, Li, Ouyang, Zhuang, Lee, Guo, et~al.]{betker2023improving}
James Betker, Gabriel Goh, Li Jing, Tim Brooks, Jianfeng Wang, Linjie Li, Long Ouyang, Juntang Zhuang, Joyce Lee, Yufei Guo, et~al.
\newblock Improving image generation with better captions.
\newblock \emph{Computer Science. https://cdn. openai. com/papers/dall-e-3. pdf}, 2023.

\bibitem[Brooks et~al.(2023)Brooks, Holynski, and Efros]{brooks2022instructpix2pix}
Tim Brooks, Aleksander Holynski, and Alexei~A. Efros.
\newblock Instructpix2pix: Learning to follow image editing instructions.
\newblock In \emph{CVPR}, 2023.

\bibitem[Brooks et~al.(2024)Brooks, Peebles, Holmes, DePue, Guo, Jing, Schnurr, Taylor, Luhman, Luhman, Ng, Wang, and Ramesh]{videoworldsimulators2024}
Tim Brooks, Bill Peebles, Connor Holmes, Will DePue, Yufei Guo, Li Jing, David Schnurr, Joe Taylor, Troy Luhman, Eric Luhman, Clarence Ng, Ricky Wang, and Aditya Ramesh.
\newblock Video generation models as world simulators.
\newblock 2024.

\bibitem[Caesar et~al.(2021)Caesar, Kabzan, Tan, Fong, Wolff, Lang, Fletcher, Beijbom, and Omari]{caesar2021nuplan}
Holger Caesar, Juraj Kabzan, Kok~Seang Tan, Whye~Kit Fong, Eric Wolff, Alex Lang, Luke Fletcher, Oscar Beijbom, and Sammy Omari.
\newblock nuplan: A closed-loop ml-based planning benchmark for autonomous vehicles.
\newblock \emph{arXiv preprint arXiv:2106.11810}, 2021.

\bibitem[Chai et~al.(2021)Chai, Wulff, and Isola]{chai2021using}
Lucy Chai, Jonas Wulff, and Phillip Isola.
\newblock Using latent space regression to analyze and leverage compositionality in gans.
\newblock In \emph{ICLR}, 2021.

\bibitem[Gao et~al.(2024{\natexlab{a}})Gao, Chen, Xie, Hong, Li, Yeung, and Xu]{gao2023magicdrive}
Ruiyuan Gao, Kai Chen, Enze Xie, Lanqing Hong, Zhenguo Li, Dit-Yan Yeung, and Qiang Xu.
\newblock Magicdrive: Street view generation with diverse 3d geometry control.
\newblock In \emph{ICLR}, 2024{\natexlab{a}}.

\bibitem[Gao et~al.(2024{\natexlab{b}})Gao, Holynski, Henzler, Brussee, Martin-Brualla, Srinivasan, Barron, and Poole]{gao2024cat3d}
Ruiqi Gao, Aleksander Holynski, Philipp Henzler, Arthur Brussee, Ricardo Martin-Brualla, Pratul Srinivasan, Jonathan~T Barron, and Ben Poole.
\newblock Cat3d: Create anything in 3d with multi-view diffusion models.
\newblock In \emph{NeurIPS}, 2024{\natexlab{b}}.

\bibitem[Gao et~al.(2024{\natexlab{c}})Gao, Yang, Chen, Chitta, Qiu, Geiger, Zhang, and Li]{gao2024vista}
Shenyuan Gao, Jiazhi Yang, Li Chen, Kashyap Chitta, Yihang Qiu, Andreas Geiger, Jun Zhang, and Hongyang Li.
\newblock Vista: A generalizable driving world model with high fidelity and versatile controllability.
\newblock In \emph{NeurIPS}, 2024{\natexlab{c}}.

\bibitem[Gu et~al.(2023)Gu, Hu, Zhang, Chen, Wang, Wang, and Zhao]{gu2023vip3d}
Junru Gu, Chenxu Hu, Tianyuan Zhang, Xuanyao Chen, Yilun Wang, Yue Wang, and Hang Zhao.
\newblock Vip3d: End-to-end visual trajectory prediction via 3d agent queries.
\newblock In \emph{CVPR}, 2023.

\bibitem[Gulino et~al.(2024)Gulino, Fu, Luo, Tucker, Bronstein, Lu, Harb, Pan, Wang, Chen, et~al.]{gulino2024waymax}
Cole Gulino, Justin Fu, Wenjie Luo, George Tucker, Eli Bronstein, Yiren Lu, Jean Harb, Xinlei Pan, Yan Wang, Xiangyu Chen, et~al.
\newblock Waymax: An accelerated, data-driven simulator for large-scale autonomous driving research.
\newblock In \emph{NeurIPS}, 2024.

\bibitem[Gupta et~al.(2024)Gupta, Yu, Sohn, Gu, Hahn, Fei-Fei, Essa, Jiang, and Lezama]{gupta2023photorealistic}
Agrim Gupta, Lijun Yu, Kihyuk Sohn, Xiuye Gu, Meera Hahn, Li Fei-Fei, Irfan Essa, Lu Jiang, and Jos{\'e} Lezama.
\newblock Photorealistic video generation with diffusion models.
\newblock In \emph{ECCV}, 2024.

\bibitem[Hertz et~al.(2023)Hertz, Mokady, Tenenbaum, Aberman, Pritch, and Cohen-Or]{hertz2022prompt}
Amir Hertz, Ron Mokady, Jay Tenenbaum, Kfir Aberman, Yael Pritch, and Daniel Cohen-Or.
\newblock Prompt-to-prompt image editing with cross attention control.
\newblock In \emph{ICLR}, 2023.

\bibitem[Hessel et~al.(2021)Hessel, Holtzman, Forbes, Bras, and Choi]{DBLP:conf/emnlp/HesselHFBC21}
Jack Hessel, Ari Holtzman, Maxwell Forbes, Ronan~Le Bras, and Yejin Choi.
\newblock Clipscore: {A} reference-free evaluation metric for image captioning.
\newblock In \emph{EMNLP}, 2021.

\bibitem[Heusel et~al.(2017)Heusel, Ramsauer, Unterthiner, Nessler, and Hochreiter]{DBLP:conf/nips/HeuselRUNH17}
Martin Heusel, Hubert Ramsauer, Thomas Unterthiner, Bernhard Nessler, and Sepp Hochreiter.
\newblock Gans trained by a two time-scale update rule converge to a local nash equilibrium.
\newblock In \emph{NeurIPS}, 2017.

\bibitem[Ho et~al.(2020)Ho, Jain, and Abbeel]{ho2020denoising}
Jonathan Ho, Ajay Jain, and Pieter Abbeel.
\newblock Denoising diffusion probabilistic models.
\newblock \emph{arXiv preprint arxiv:2006.11239}, 2020.

\bibitem[Ho et~al.(2022)Ho, Chan, Saharia, Whang, Gao, Gritsenko, Kingma, Poole, Norouzi, Fleet, et~al.]{ho2022imagen}
Jonathan Ho, William Chan, Chitwan Saharia, Jay Whang, Ruiqi Gao, Alexey Gritsenko, Diederik~P Kingma, Ben Poole, Mohammad Norouzi, David~J Fleet, et~al.
\newblock Imagen video: High definition video generation with diffusion models.
\newblock \emph{arXiv preprint arXiv:2210.02303}, 2022.

\bibitem[Hu et~al.(2023{\natexlab{a}})Hu, Russell, Yeo, Murez, Fedoseev, Kendall, Shotton, and Corrado]{hu2023gaia1generativeworldmodel}
Anthony Hu, Lloyd Russell, Hudson Yeo, Zak Murez, George Fedoseev, Alex Kendall, Jamie Shotton, and Gianluca Corrado.
\newblock Gaia-1: A generative world model for autonomous driving, 2023{\natexlab{a}}.

\bibitem[Hu et~al.(2022)Hu, Chen, Wu, Li, Yan, and Tao]{hu2022st}
Shengchao Hu, Li Chen, Penghao Wu, Hongyang Li, Junchi Yan, and Dacheng Tao.
\newblock St-p3: End-to-end vision-based autonomous driving via spatial-temporal feature learning.
\newblock In \emph{ECCV}, 2022.

\bibitem[Hu et~al.(2023{\natexlab{b}})Hu, Yang, Chen, Li, Sima, Zhu, Chai, Du, Lin, Wang, et~al.]{hu2023planning}
Yihan Hu, Jiazhi Yang, Li Chen, Keyu Li, Chonghao Sima, Xizhou Zhu, Siqi Chai, Senyao Du, Tianwei Lin, Wenhai Wang, et~al.
\newblock Planning-oriented autonomous driving.
\newblock In \emph{CVPR}, 2023{\natexlab{b}}.

\bibitem[Hwang et~al.(2024)Hwang, Xu, Lin, Hung, Ji, Choi, Huang, He, Covington, Sapp, Guo, Anguelov, and Tan]{hwang2024emmaendtoendmultimodalmodel}
Jyh-Jing Hwang, Runsheng Xu, Hubert Lin, Wei-Chih Hung, Jingwei Ji, Kristy Choi, Di Huang, Tong He, Paul Covington, Benjamin Sapp, James Guo, Dragomir Anguelov, and Mingxing Tan.
\newblock Emma: End-to-end multimodal model for autonomous driving, 2024.

\bibitem[Jaegle et~al.(2022)Jaegle, Borgeaud, Alayrac, Doersch, Ionescu, Ding, Koppula, Zoran, Brock, Shelhamer, Henaff, Botvinick, Zisserman, Vinyals, and Carreira]{jaegle2022perceiver}
Andrew Jaegle, Sebastian Borgeaud, Jean-Baptiste Alayrac, Carl Doersch, Catalin Ionescu, David Ding, Skanda Koppula, Daniel Zoran, Andrew Brock, Evan Shelhamer, Olivier~J Henaff, Matthew Botvinick, Andrew Zisserman, Oriol Vinyals, and Joao Carreira.
\newblock Perceiver {IO}: A general architecture for structured inputs \& outputs.
\newblock In \emph{ICLR}, 2022.

\bibitem[Karras et~al.(2019)Karras, Laine, and Aila]{karras2019style}
Tero Karras, Samuli Laine, and Timo Aila.
\newblock A style-based generator architecture for generative adversarial networks.
\newblock In \emph{CVPR}, 2019.

\bibitem[Karras et~al.(2020)Karras, Laine, Aittala, Hellsten, Lehtinen, and Aila]{karras2020analyzing}
Tero Karras, Samuli Laine, Miika Aittala, Janne Hellsten, Jaakko Lehtinen, and Timo Aila.
\newblock Analyzing and improving the image quality of stylegan.
\newblock In \emph{CVPR}, 2020.

\bibitem[Kerbl et~al.(2023)Kerbl, Kopanas, Leimk{\"u}hler, and Drettakis]{kerbl20233d}
Bernhard Kerbl, Georgios Kopanas, Thomas Leimk{\"u}hler, and George Drettakis.
\newblock 3d gaussian splatting for real-time radiance field rendering.
\newblock \emph{ACM TOG}, 42\penalty0 (4):\penalty0 139--1, 2023.

\bibitem[Kong et~al.(2020)Kong, Ping, Huang, Zhao, and Catanzaro]{kong2020diffwave}
Zhifeng Kong, Wei Ping, Jiaji Huang, Kexin Zhao, and Bryan Catanzaro.
\newblock Diffwave: A versatile diffusion model for audio synthesis.
\newblock \emph{arXiv preprint arXiv:2009.09761}, 2020.

\bibitem[Li et~al.(2023{\natexlab{a}})Li, Ling, Kar, Acuna, Kim, Kreis, Torralba, and Fidler]{li2023dreamteacher}
Daiqing Li, Huan Ling, Amlan Kar, David Acuna, Seung~Wook Kim, Karsten Kreis, Antonio Torralba, and Sanja Fidler.
\newblock Dreamteacher: Pretraining image backbones with deep generative models.
\newblock In \emph{ICCV}, 2023{\natexlab{a}}.

\bibitem[Li et~al.(2022)Li, Thickstun, Gulrajani, Liang, and Hashimoto]{li2022diffusion}
Xiang Li, John Thickstun, Ishaan Gulrajani, Percy~S Liang, and Tatsunori~B Hashimoto.
\newblock Diffusion-lm improves controllable text generation.
\newblock In \emph{NeurIPS}, 2022.

\bibitem[Li et~al.(2023{\natexlab{b}})Li, Zhang, and Ye]{li2023drivingdiffusion}
Xiaofan Li, Yifu Zhang, and Xiaoqing Ye.
\newblock Drivingdiffusion: Layout-guided multi-view driving scene video generation with latent diffusion model.
\newblock \emph{arXiv preprint arXiv:2310.07771}, 2023{\natexlab{b}}.

\bibitem[Liang et~al.(2024)Liang, Yan, Chen, Zhou, Yan, Zhong, and Zou]{liang2024driveeditor}
Yiyuan Liang, Zhiying Yan, Liqun Chen, Jiahuan Zhou, Luxin Yan, Sheng Zhong, and Xu Zou.
\newblock Driveeditor: A unified 3d information-guided framework for controllable object editing in driving scenes, 2024.

\bibitem[Lugmayr et~al.(2022)Lugmayr, Danelljan, Romero, Yu, Timofte, and Van~Gool]{lugmayr2022repaint}
Andreas Lugmayr, Martin Danelljan, Andres Romero, Fisher Yu, Radu Timofte, and Luc Van~Gool.
\newblock Repaint: Inpainting using denoising diffusion probabilistic models.
\newblock In \emph{CVPR}, 2022.

\bibitem[Ma et~al.(2024)Ma, Zhou, Tang, Zhang, Han, Jiang, Zhan, Jia, Lang, Sun, Lin, and Yu]{ma2024unleashinggeneralizationendtoendautonomous}
Enhui Ma, Lijun Zhou, Tao Tang, Zhan Zhang, Dong Han, Junpeng Jiang, Kun Zhan, Peng Jia, Xianpeng Lang, Haiyang Sun, Di Lin, and Kaicheng Yu.
\newblock Unleashing generalization of end-to-end autonomous driving with controllable long video generation, 2024.

\bibitem[Mann et~al.(2020)Mann, Ryder, Subbiah, Kaplan, Dhariwal, Neelakantan, Shyam, Sastry, Askell, Agarwal, et~al.]{mann2020language}
Ben Mann, N Ryder, M Subbiah, J Kaplan, P Dhariwal, A Neelakantan, P Shyam, G Sastry, A Askell, S Agarwal, et~al.
\newblock Language models are few-shot learners.
\newblock In \emph{NeurIPS}, 2020.

\bibitem[Meng et~al.(2022)Meng, He, Song, Song, Wu, Zhu, and Ermon]{meng2021sdedit}
Chenlin Meng, Yutong He, Yang Song, Jiaming Song, Jiajun Wu, Jun-Yan Zhu, and Stefano Ermon.
\newblock Sdedit: Guided image synthesis and editing with stochastic differential equations.
\newblock In \emph{ICLR}, 2022.

\bibitem[Mildenhall et~al.(2020)Mildenhall, Srinivasan, Tancik, Barron, Ramamoorthi, and Ng]{mildenhall2020nerf}
Ben Mildenhall, Pratul~P Srinivasan, Matthew Tancik, Jonathan~T Barron, Ravi Ramamoorthi, and Ren Ng.
\newblock Nerf: Representing scenes as neural radiance fields for view synthesis.
\newblock In \emph{ECCV}, 2020.

\bibitem[Montali et~al.(2024)Montali, Lambert, Mougin, Kuefler, Rhinehart, Li, Gulino, Emrich, Yang, Whiteson, et~al.]{montali2024waymo}
Nico Montali, John Lambert, Paul Mougin, Alex Kuefler, Nicholas Rhinehart, Michelle Li, Cole Gulino, Tristan Emrich, Zoey Yang, Shimon Whiteson, et~al.
\newblock The waymo open sim agents challenge.
\newblock In \emph{NeurIPS}, 2024.

\bibitem[Nichol et~al.(2022)Nichol, Dhariwal, Ramesh, Shyam, Mishkin, McGrew, Sutskever, and Chen]{nichol2022glide}
Alex Nichol, Prafulla Dhariwal, Aditya Ramesh, Pranav Shyam, Pamela Mishkin, Bob McGrew, Ilya Sutskever, and Mark Chen.
\newblock Glide: Towards photorealistic image generation and editing with text-guided diffusion models.
\newblock In \emph{ICML}, 2022.

\bibitem[Radford et~al.(2021)Radford, Kim, Hallacy, Ramesh, Goh, Agarwal, Sastry, Askell, Mishkin, Clark, et~al.]{radford2021learning}
Alec Radford, Jong~Wook Kim, Chris Hallacy, Aditya Ramesh, Gabriel Goh, Sandhini Agarwal, Girish Sastry, Amanda Askell, Pamela Mishkin, Jack Clark, et~al.
\newblock Learning transferable visual models from natural language supervision.
\newblock In \emph{ICML}, 2021.

\bibitem[Ramesh et~al.(2022)Ramesh, Dhariwal, Nichol, Chu, and Chen]{ramesh2022hierarchical}
Aditya Ramesh, Prafulla Dhariwal, Alex Nichol, Casey Chu, and Mark Chen.
\newblock Hierarchical text-conditional image generation with clip latents.
\newblock \emph{arXiv preprint arXiv:2204.06125}, 2022.

\bibitem[Richardson et~al.(2021)Richardson, Alaluf, Patashnik, Nitzan, Azar, Shapiro, and Cohen-Or]{richardson2021encoding}
Elad Richardson, Yuval Alaluf, Or Patashnik, Yotam Nitzan, Yaniv Azar, Stav Shapiro, and Daniel Cohen-Or.
\newblock Encoding in style: a stylegan encoder for image-to-image translation.
\newblock In \emph{CVPR}, 2021.

\bibitem[Rombach et~al.(2022)Rombach, Blattmann, Lorenz, Esser, and Ommer]{rombach2022high}
Robin Rombach, Andreas Blattmann, Dominik Lorenz, Patrick Esser, and Bj{\"o}rn Ommer.
\newblock High-resolution image synthesis with latent diffusion models.
\newblock In \emph{CVPR}, 2022.

\bibitem[Saharia et~al.(2022)Saharia, Chan, Saxena, Li, Whang, Denton, Ghasemipour, Gontijo~Lopes, Karagol~Ayan, Salimans, et~al.]{saharia2022photorealistic}
Chitwan Saharia, William Chan, Saurabh Saxena, Lala Li, Jay Whang, Emily~L Denton, Kamyar Ghasemipour, Raphael Gontijo~Lopes, Burcu Karagol~Ayan, Tim Salimans, et~al.
\newblock Photorealistic text-to-image diffusion models with deep language understanding.
\newblock In \emph{NeurIPS}, 2022.

\bibitem[Sohl-Dickstein et~al.(2015)Sohl-Dickstein, Weiss, Maheswaranathan, and Ganguli]{sohl2015deep}
Jascha Sohl-Dickstein, Eric Weiss, Niru Maheswaranathan, and Surya Ganguli.
\newblock Deep unsupervised learning using nonequilibrium thermodynamics.
\newblock In \emph{ICML}, 2015.

\bibitem[Song et~al.(2020)Song, Meng, and Ermon]{song2020denoising}
Jiaming Song, Chenlin Meng, and Stefano Ermon.
\newblock Denoising diffusion implicit models.
\newblock \emph{arXiv:2010.02502}, 2020.

\bibitem[Song and Ermon(2019)]{song2019generative}
Yang Song and Stefano Ermon.
\newblock Generative modeling by estimating gradients of the data distribution.
\newblock In \emph{NeurIPS}, 2019.

\bibitem[Tov et~al.(2021)Tov, Alaluf, Nitzan, Patashnik, and Cohen-Or]{tov2021designing}
Omer Tov, Yuval Alaluf, Yotam Nitzan, Or Patashnik, and Daniel Cohen-Or.
\newblock Designing an encoder for stylegan image manipulation.
\newblock \emph{ACM TOG}, 2021.

\bibitem[Wang et~al.(2023)Wang, Zhu, Huang, Chen, Zhu, and Lu]{wang2023drivedreamerrealworlddrivenworldmodels}
Xiaofeng Wang, Zheng Zhu, Guan Huang, Xinze Chen, Jiagang Zhu, and Jiwen Lu.
\newblock Drivedreamer: Towards real-world-driven world models for autonomous driving, 2023.

\bibitem[Wang et~al.(2024)Wang, He, Fan, Li, Chen, and Zhang]{wang2024driving}
Yuqi Wang, Jiawei He, Lue Fan, Hongxin Li, Yuntao Chen, and Zhaoxiang Zhang.
\newblock Driving into the future: Multiview visual forecasting and planning with world model for autonomous driving.
\newblock In \emph{CVPR}, 2024.

\bibitem[Wei et~al.(2024)Wei, Wang, Lu, Xu, Liu, Zhao, Chen, and Wang]{wei2024editable}
Yuxi Wei, Zi Wang, Yifan Lu, Chenxin Xu, Changxing Liu, Hao Zhao, Siheng Chen, and Yanfeng Wang.
\newblock Editable scene simulation for autonomous driving via collaborative llm-agents.
\newblock In \emph{Proceedings of the IEEE/CVF Conference on Computer Vision and Pattern Recognition (CVPR)}, 2024.

\bibitem[Wen et~al.(2024)Wen, Zhao, Liu, Jia, Wang, Luo, Zhang, Wang, Sun, and Zhang]{wen2024panacea}
Yuqing Wen, Yucheng Zhao, Yingfei Liu, Fan Jia, Yanhui Wang, Chong Luo, Chi Zhang, Tiancai Wang, Xiaoyan Sun, and Xiangyu Zhang.
\newblock Panacea: Panoramic and controllable video generation for autonomous driving.
\newblock In \emph{CVPR}, 2024.

\bibitem[Wu et~al.(2023{\natexlab{a}})Wu, Zhao, Chen, Gu, Zhao, He, Zhou, Shou, and Shen]{wu2023datasetdm}
Weijia Wu, Yuzhong Zhao, Hao Chen, Yuchao Gu, Rui Zhao, Yefei He, Hong Zhou, Mike~Zheng Shou, and Chunhua Shen.
\newblock Datasetdm: Synthesizing data with perception annotations using diffusion models.
\newblock 2023{\natexlab{a}}.

\bibitem[Wu et~al.(2023{\natexlab{b}})Wu, Zhao, Shou, Zhou, and Shen]{wu2023diffumask}
Weijia Wu, Yuzhong Zhao, Mike~Zheng Shou, Hong Zhou, and Chunhua Shen.
\newblock Diffumask: Synthesizing images with pixel-level annotations for semantic segmentation using diffusion models.
\newblock In \emph{ICCV}, 2023{\natexlab{b}}.

\bibitem[Yang et~al.(2024)Yang, Gao, Qiu, Chen, Li, Dai, Chitta, Wu, Zeng, Luo, Zhang, Geiger, Qiao, and Li]{yang2024genad}
Jiazhi Yang, Shenyuan Gao, Yihang Qiu, Li Chen, Tianyu Li, Bo Dai, Kashyap Chitta, Penghao Wu, Jia Zeng, Ping Luo, Jun Zhang, Andreas Geiger, Yu Qiao, and Hongyang Li.
\newblock {Generalized Predictive Model for Autonomous Driving}.
\newblock In \emph{CVPR}, 2024.

\bibitem[Yang et~al.(2023)Yang, Chen, Wang, Manivasagam, Ma, Yang, and Urtasun]{yang2023unisim}
Ze Yang, Yun Chen, Jingkang Wang, Sivabalan Manivasagam, Wei-Chiu Ma, Anqi~Joyce Yang, and Raquel Urtasun.
\newblock Unisim: A neural closed-loop sensor simulator.
\newblock In \emph{CVPR}, 2023.

\bibitem[Zhang et~al.(2018)Zhang, Isola, Efros, Shechtman, and Wang]{zhang2018unreasonable}
Richard Zhang, Phillip Isola, Alexei~A Efros, Eli Shechtman, and Oliver Wang.
\newblock The unreasonable effectiveness of deep features as a perceptual metric.
\newblock In \emph{CVPR}, 2018.

\bibitem[Zhang et~al.(2021)Zhang, Ling, Gao, Yin, Lafleche, Barriuso, Torralba, and Fidler]{zhang2021datasetgan}
Yuxuan Zhang, Huan Ling, Jun Gao, Kangxue Yin, Jean-Francois Lafleche, Adela Barriuso, Antonio Torralba, and Sanja Fidler.
\newblock Datasetgan: Efficient labeled data factory with minimal human effort.
\newblock In \emph{CVPR}, 2021.

\bibitem[Zhao et~al.(2024)Zhao, Wang, Zhu, Chen, Huang, Bao, and Wang]{zhao2024drive}
Guosheng Zhao, Xiaofeng Wang, Zheng Zhu, Xinze Chen, Guan Huang, Xiaoyi Bao, and Xingang Wang.
\newblock Drivedreamer-2: Llm-enhanced world models for diverse driving video generation.
\newblock \emph{arXiv preprint arXiv:2403.06845}, 2024.

\end{thebibliography}
}

\clearpage
\setcounter{page}{1}
\maketitle
\renewcommand{\thesection}{\Alph{section}}
\setcounter{section}{0}

\section{More Technical Details}

\begin{figure*}[t]
    \includegraphics[width=0.95\linewidth]{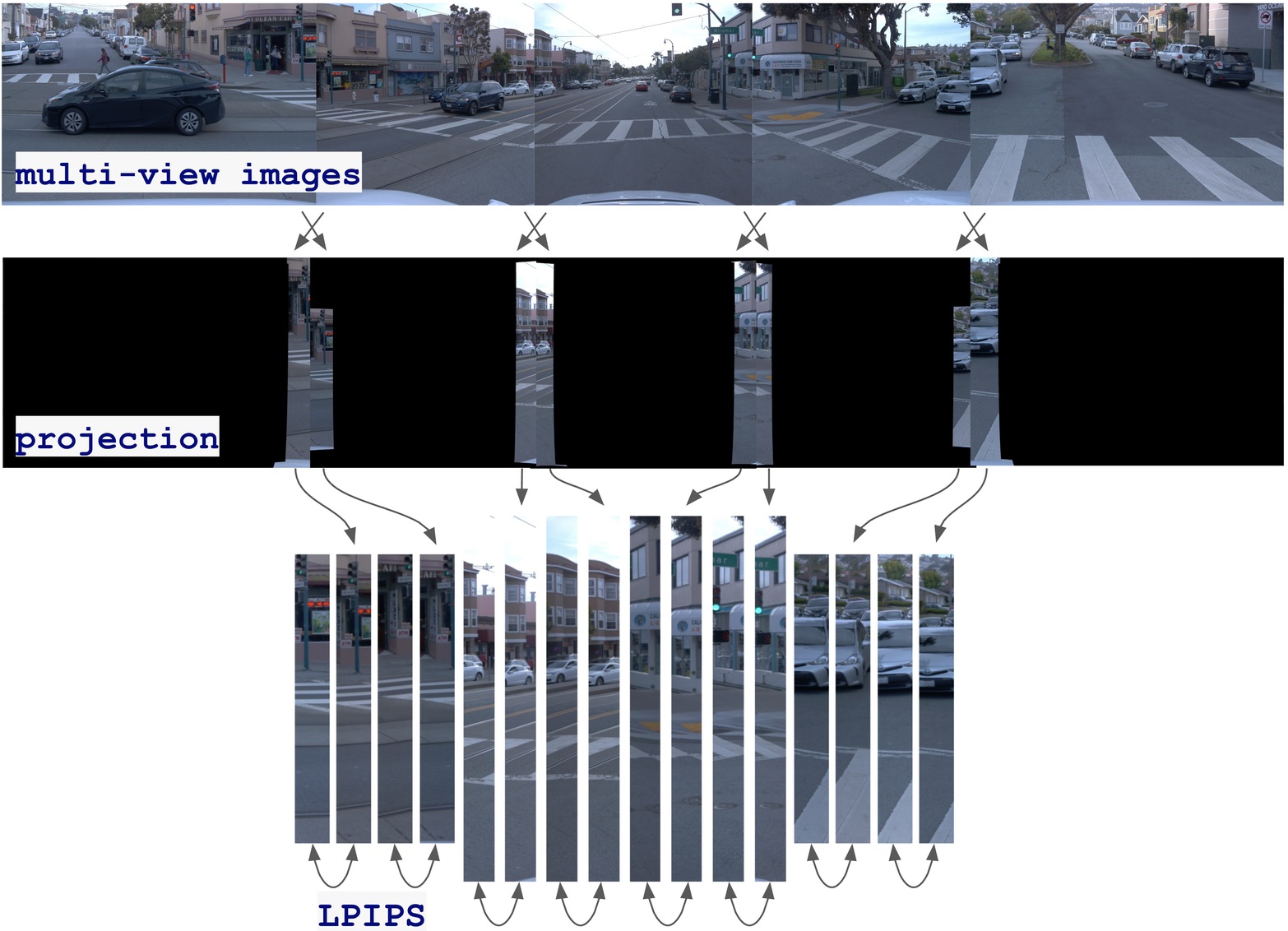}
    
    \caption{\textbf{Illustration of our multi-view consistency metric.} Starting from a multi-view image, we project each view into the two adjacent views. We are then able to define two pairs of image patches for each overlapping region and compare them using LPIPS to evaluate their consistency. For simplicity, we show an illustrative example with only 5 cameras.}
    \label{fig:consistency_metric}
\end{figure*}

\subsection{SceneCrafter Models}
We trained a total of three diffusion models: the SceneCrafter teacher model for global edits (Teacher-G), the SceneCrafter teacher model for local edits (Teacher-L), and the SceneCrafter student model (Student). \tabref{model_details} provides an overview of the experimental settings for each of these models.

\subsection{Inference Time}
Our method takes approximately 10 seconds to generate or edit 8 multi-view images with $512\times512$ size each on a single A100 GPU. For larger-scale operations, we run it on a cluster of 128 A100 GPUs, achieving an amortized inference time of just 0.1 seconds.

\subsection{Multi-View Consistency}
\begin{figure*}[t]
    \centering
    
\resizebox{\linewidth}{!}{
\begin{tikzpicture}

\def\imagewidth{14.4cm}
\def\imageheight{1.8cm}
\def\verticalgap{-0.2cm}

\node (image1) {\includegraphics[width=\imagewidth, height=\imageheight]{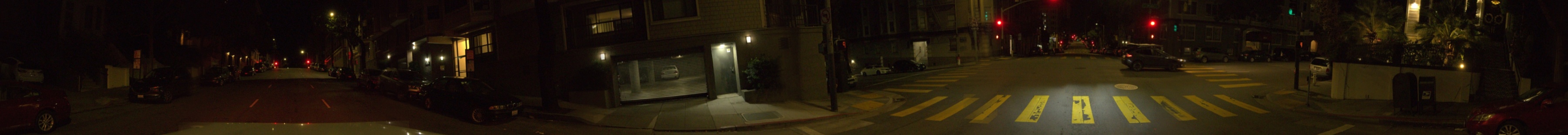}}; 
\node[anchor=east] at (image1.west) {Real}; 

\node[below=\verticalgap of image1] (image1_gen) {\includegraphics[width=\imagewidth, height=\imageheight]{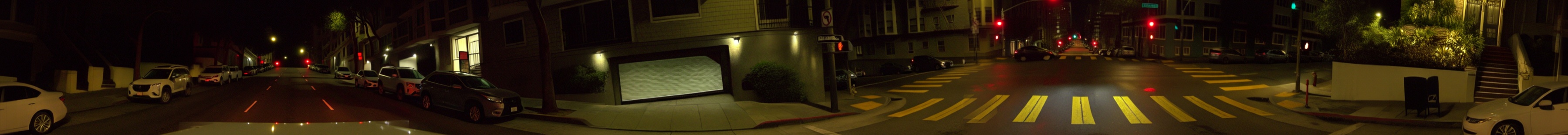}}; 
\node[anchor=east] at (image1_gen.west) {Ours}; 

\node[below=\verticalgap of image1_gen] (image2) {\includegraphics[width=\imagewidth, height=\imageheight]{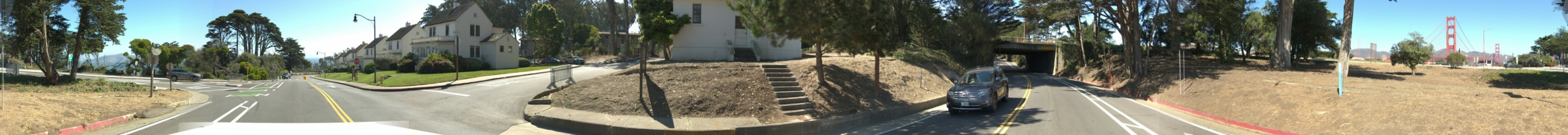}}; 
\node[anchor=east] at (image2.west) {Real}; 

\node[below=\verticalgap of image2] (image2_gen) {\includegraphics[width=\imagewidth, height=\imageheight]{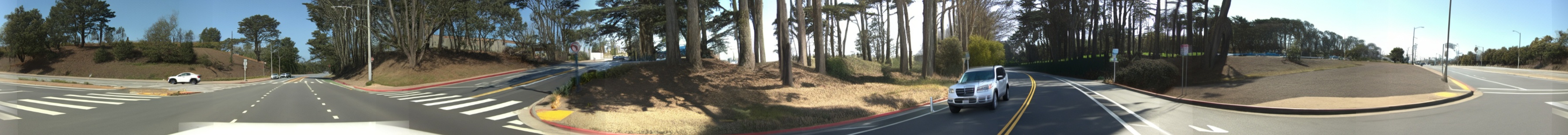}}; 
\node[anchor=east] at (image2_gen.west) {Ours}; 

\node[below=\verticalgap of image2_gen] (image3) {\includegraphics[width=\imagewidth, height=\imageheight]{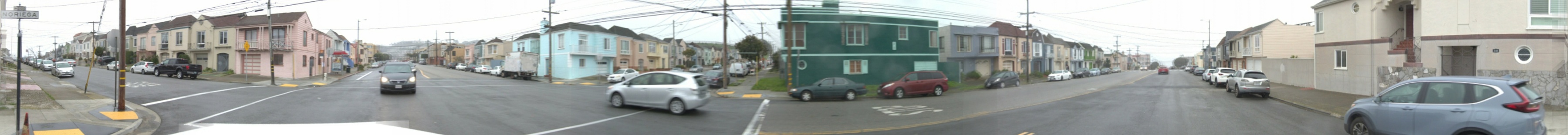}}; 
\node[anchor=east] at (image3.west) {Real}; 

\node[below=\verticalgap of image3] (image3_gen) {\includegraphics[width=\imagewidth, height=\imageheight]{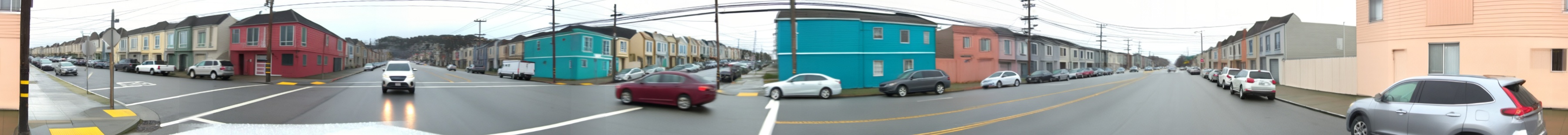}}; 
\node[anchor=east] at (image3_gen.west) {Ours}; 

\node[below=\verticalgap of image3_gen] (image4) {\includegraphics[width=\imagewidth, height=\imageheight]{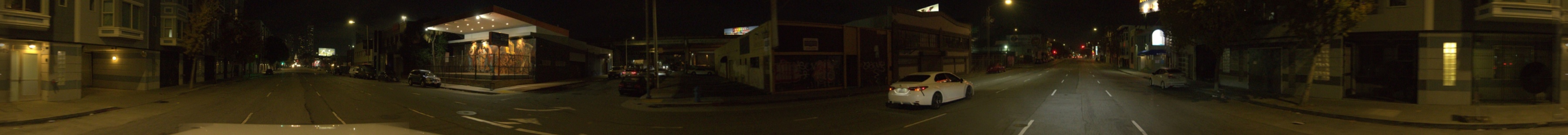}}; 
\node[anchor=east] at (image4.west) {Real}; 

\node[below=\verticalgap of image4] (image4_gen) {\includegraphics[width=\imagewidth, height=\imageheight]{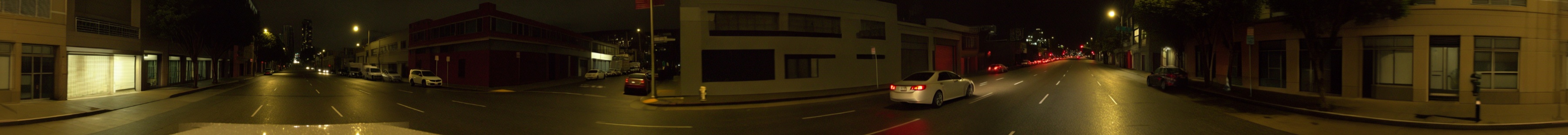}}; 
\node[anchor=east] at (image4_gen.west) {Ours};

\end{tikzpicture}
}
 \caption{\textbf{Panoramic Images.} Here we show panoramic images generated by stitching individual camera views into one 360$^{\circ}$ surround view. Both real imagery as well as our method with conditioning produce no visible seams or inconsistencies beyond projection error.}
    \label{fig:consistency_panos}
\end{figure*}
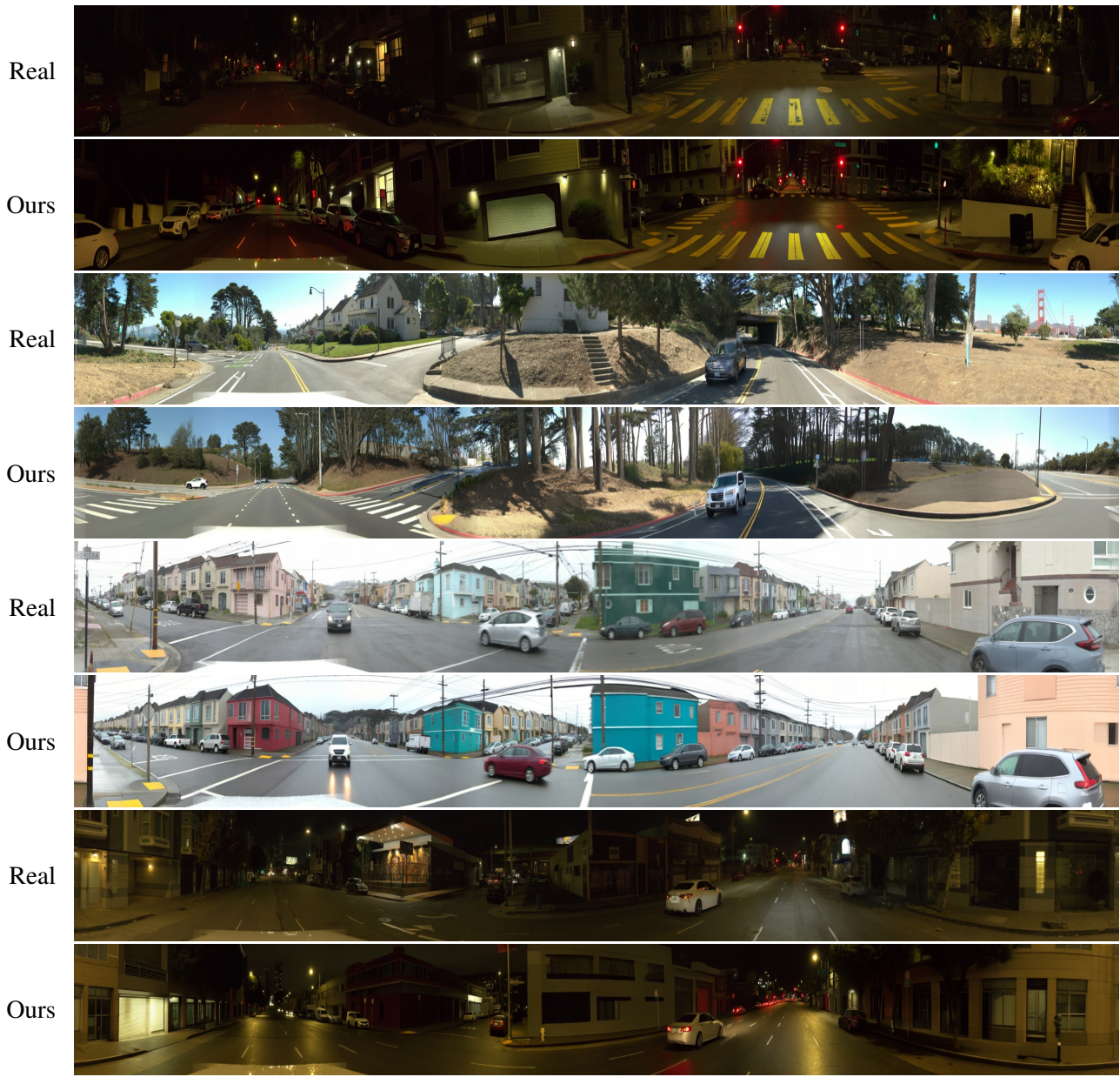

Our model is trained on 8 cameras whose fields of view overlap by several degrees between each neighboring pair. Therefore our metric is evaluated on a total of 16 image pairs per frame: 8 pairs by projecting from $C_i$ to $C_{i+1}$, and 8 pairs by projecting from $C_{i+1}$ to $C_i$. We illustrate this process in \figref{consistency_metric}. For the projection, we utilize known camera intrinsic and extrinsic parameters, and assume a fixed baseline distance. While this is an approximation, the LPIPS metric is well documented to be sufficiently robust to minor spatial shifts and distortions~\cite{zhang2018unreasonable}. \figref{consistency_panos} shows stitched 360$^{\circ}$ panoramic images of real and generated images with no visible seams or inconsistencies beyond projection error.

\begin{table}[t]
\centering
\resizebox{0.99\linewidth}{!}{
\begin{tabular}{lccc}
\toprule
\multicolumn{1}{l}{} & Teacher-G  &  Teacher-L  & Student  \\
\midrule
\textbf{\textit{Condition}}\\
Time of day & \cmark & \cmark & \cmark    \\
Weather & \cmark & \cmark & \cmark    \\
Agent boxes & \cmark & \xmark & \cmark    \\
HD maps & \cmark & \cmark & \cmark     \\
Foreground masks & \xmark & \cmark & \xmark     \\
Raymaps & \cmark & \cmark & \cmark     \\
Source image & \xmark & \xmark & \cmark     \\
\midrule
\textbf{\textit{Training}}\\
Batch size & 128 & 128 & 128    \\
Learning rate &  0.0005 &  0.0005 &  0.0005    \\
Training steps &  100k &  100k &  200k   \\
Training data &  Real &  Real &  Synthetic   \\
Masked training & \xmark & \cmark & \xmark \\
Pretrained model &  CAT3D~\cite{gao2024cat3d} &  CAT3D~\cite{gao2024cat3d} &   Teacher-G    \\
\midrule
\textbf{\textit{Diffusion}}\\
Denoising Steps & 50 & 50 & 50    \\
Noise schedule & Linear & Linear & Linear    \\
EMA & \cmark & \cmark & \cmark     \\
Sampler & DDIM~\cite{song2020denoising} & DDIM~\cite{song2020denoising} & DDIM~\cite{song2020denoising}     \\
\textit{z}-shape & 64$\times$64$\times$8 & 64$\times$64$\times$8 & 64$\times$64$\times$16    \\
\midrule
\textbf{\textit{Misc}}\\
Generation ability & \cmark & \cmark & \cmark    \\
Editing ability & \xmark & \xmark & \cmark    \\
\bottomrule
\end{tabular}
}
\caption{{\bf Experimental settings for the SceneCrafter model family.} \textit{Teacher-G} refers to the SceneCrafter teacher model for global editing data generation, \textit{Teacher-T} refers to the SceneCrafter teacher model for local editing data generation, and \textit{Student} represents the SceneCrafter student model for scene editing.
}\label{tab:model_details}
\end{table}

\section{More Experiment Results}
\begin{figure*}[t]
    \centering

\resizebox{\linewidth}{!}{
\begin{tikzpicture}

\def\imagewidth{14.4cm}
\def\imageheight{1.8cm}
\def\verticalgap{-0.25cm} 

\node (image1) {\includegraphics[width=\imagewidth, height=\imageheight]{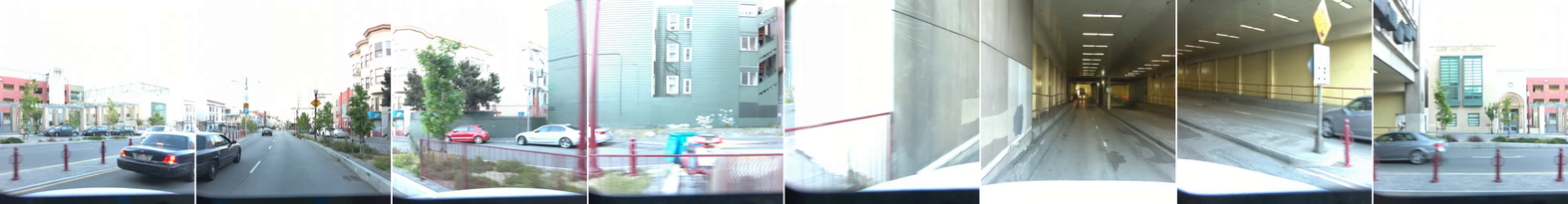}}; 
\node[anchor=east] at (image1.west) {Source}; 

\node[below=\verticalgap of image1] (image2) {\includegraphics[width=\imagewidth, height=\imageheight]{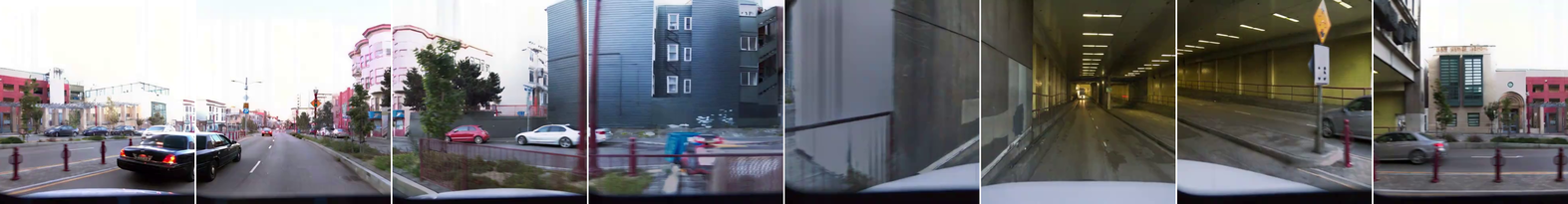}}; 
\node[anchor=east, align=right] at (image2.west) {Target \\ 7:00PM}; 

\node[below=\verticalgap of image2] (image3) {\includegraphics[width=\imagewidth, height=\imageheight]{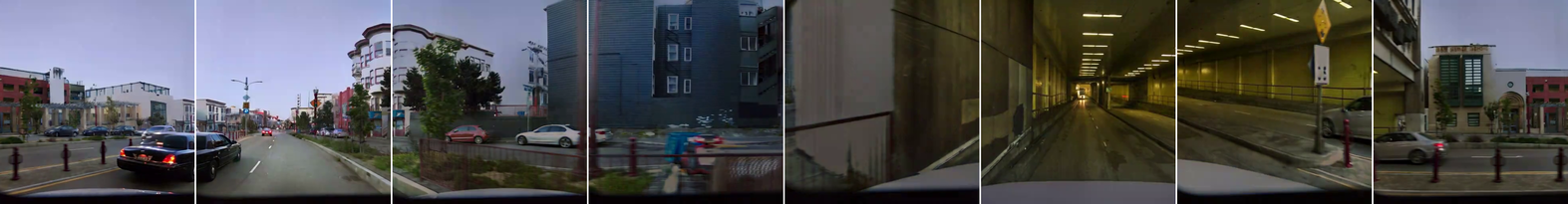}}; 
\node[anchor=east, align=right] at (image3.west) {Target \\ 7:10PM}; 

\node[below=\verticalgap of image3] (image4) {\includegraphics[width=\imagewidth, height=\imageheight]{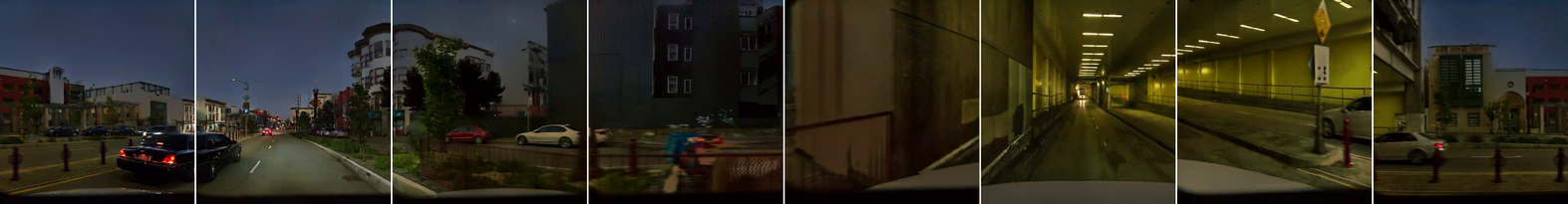}}; 
\node[anchor=east, align=right] at (image4.west) {Target \\ 7:20PM};

\node[below=\verticalgap of image4] (image5) {\includegraphics[width=\imagewidth, height=\imageheight]{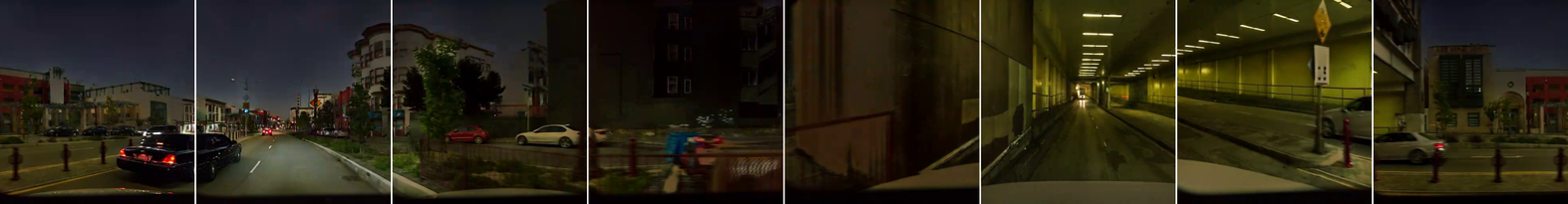}}; 
\node[anchor=east, align=right] at (image5.west) {Target \\ 7:30PM};

\node[below=\verticalgap of image5] (image6) {\includegraphics[width=\imagewidth, height=\imageheight]{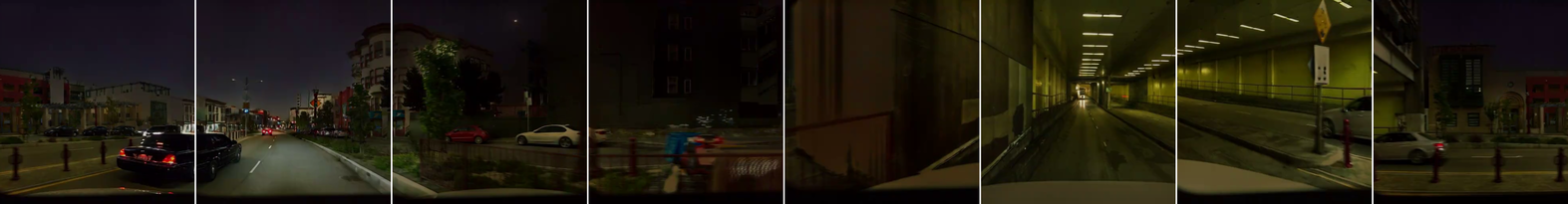}}; 
\node[anchor=east, align=right] at (image6.west) {Target \\ 7:40PM};

\node[below=\verticalgap of image6] (image7) {\includegraphics[width=\imagewidth, height=\imageheight]{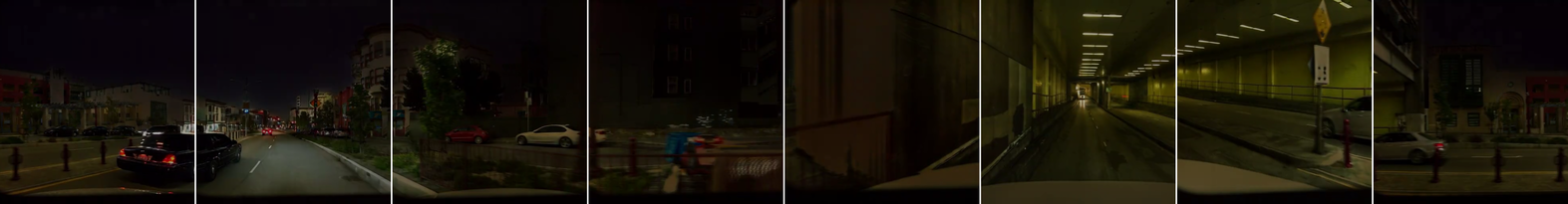}}; 
\node[anchor=east, align=right] at (image7.west) {Target \\ 7:50PM};

\node[below=\verticalgap of image7] (image8) {\includegraphics[width=\imagewidth, height=\imageheight]{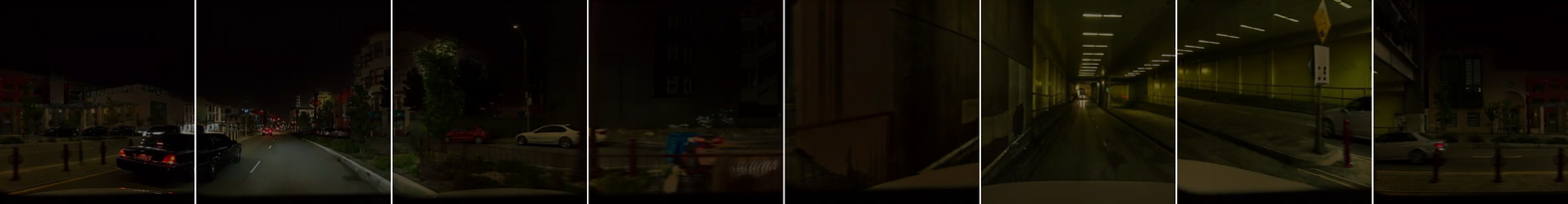}}; 
\node[anchor=east, align=right] at (image8.west) {Target \\ 8:00PM};
\end{tikzpicture}
}

\caption{
\textbf{More Visualizations on time of day editing.} We uniformly sample times between 7 PM and 8 PM to edit the source images (first row), effectively simulating day-to-night transitions.
}
\label{fig:time_edit1}
\end{figure*}

\begin{figure*}[t]
    \centering

\resizebox{\linewidth}{!}{
\begin{tikzpicture}

\def\imagewidth{14.4cm}
\def\imageheight{1.8cm}
\def\verticalgap{-0.25cm} 

\node (image1) {\includegraphics[width=\imagewidth, height=\imageheight]{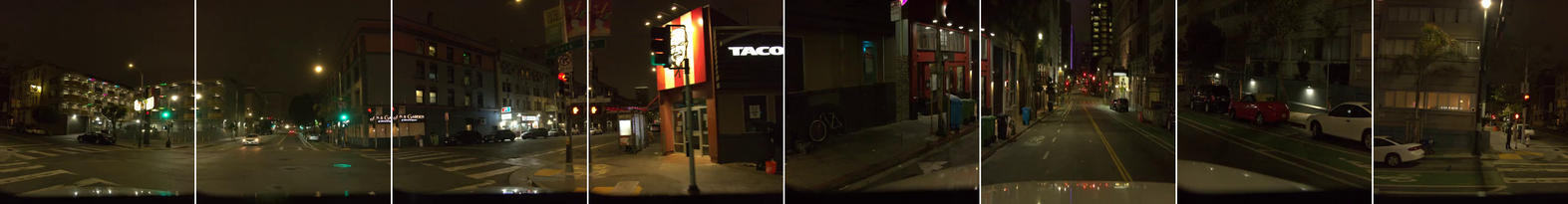}}; 
\node[anchor=east] at (image1.west) {Source}; 

\node[below=\verticalgap of image1] (image2) {\includegraphics[width=\imagewidth, height=\imageheight]{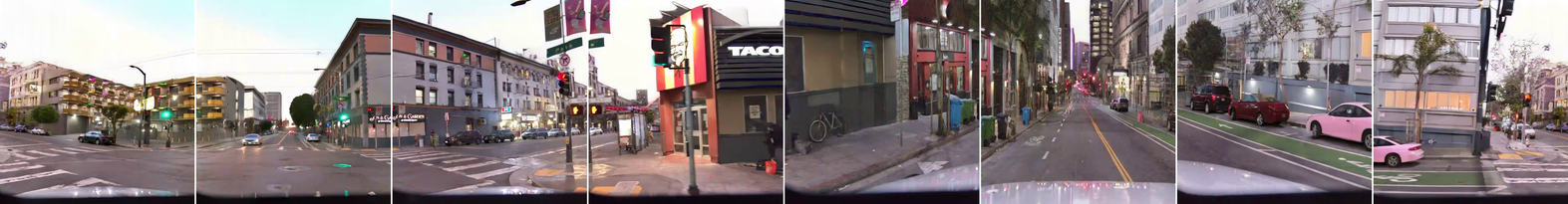}}; 
\node[anchor=east, align=right] at (image2.west) {Target \\ 7:00PM}; 

\node[below=\verticalgap of image2] (image3) {\includegraphics[width=\imagewidth, height=\imageheight]{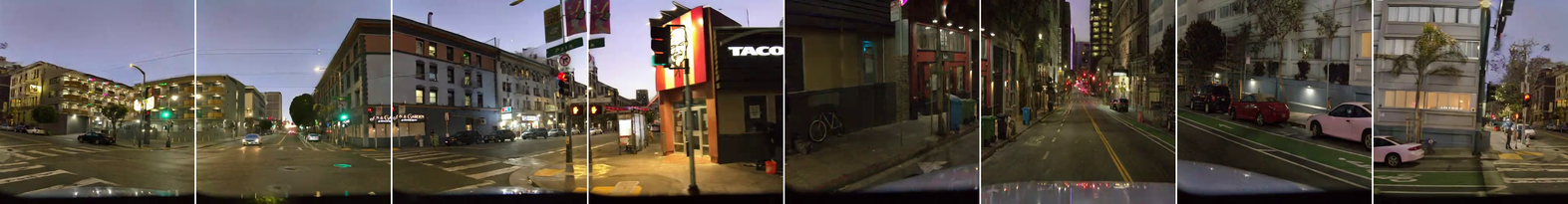}}; 
\node[anchor=east, align=right] at (image3.west) {Target \\ 7:10PM}; 

\node[below=\verticalgap of image3] (image4) {\includegraphics[width=\imagewidth, height=\imageheight]{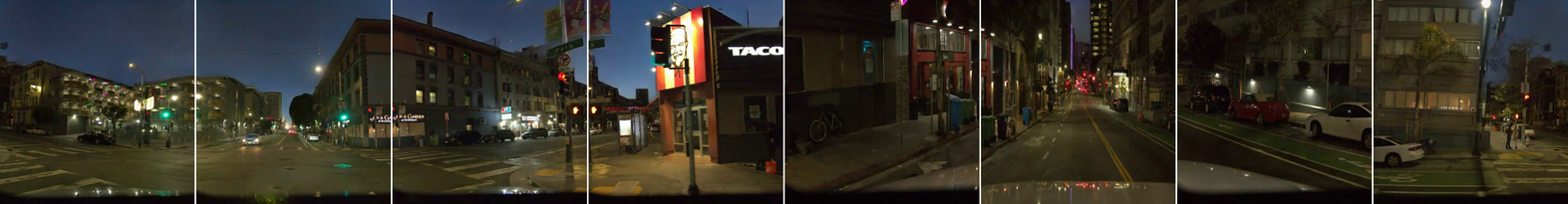}}; 
\node[anchor=east, align=right] at (image4.west) {Target \\ 7:20PM};

\node[below=\verticalgap of image4] (image5) {\includegraphics[width=\imagewidth, height=\imageheight]{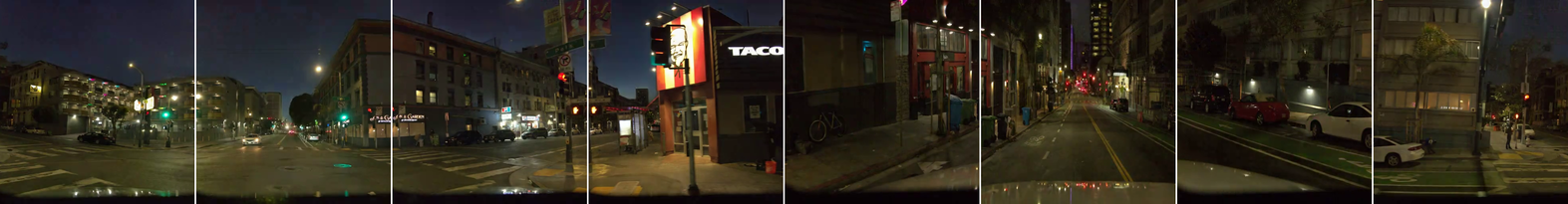}}; 
\node[anchor=east, align=right] at (image5.west) {Target \\ 7:30PM};

\node[below=\verticalgap of image5] (image6) {\includegraphics[width=\imagewidth, height=\imageheight]{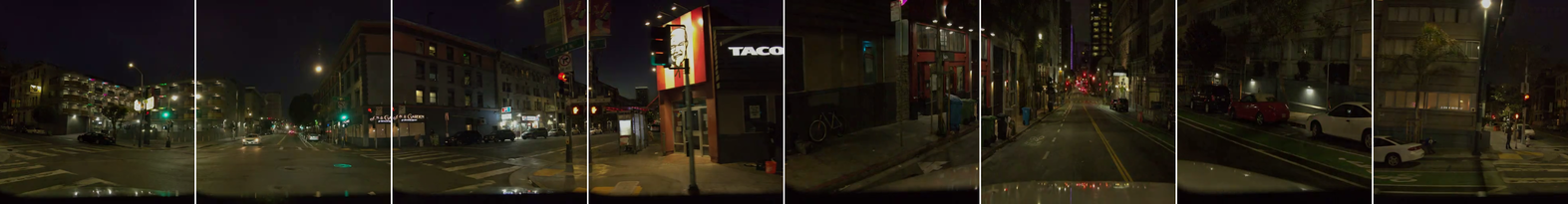}}; 
\node[anchor=east, align=right] at (image6.west) {Target \\ 7:40PM};

\node[below=\verticalgap of image6] (image7) {\includegraphics[width=\imagewidth, height=\imageheight]{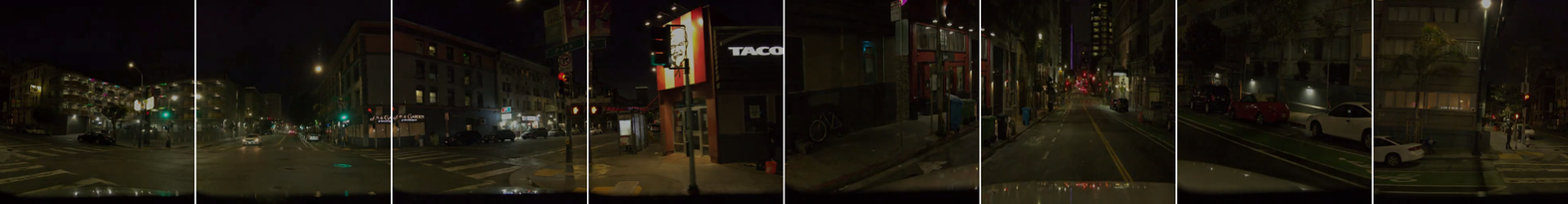}}; 
\node[anchor=east, align=right] at (image7.west) {Target \\ 7:50PM};

\node[below=\verticalgap of image7] (image8) {\includegraphics[width=\imagewidth, height=\imageheight]{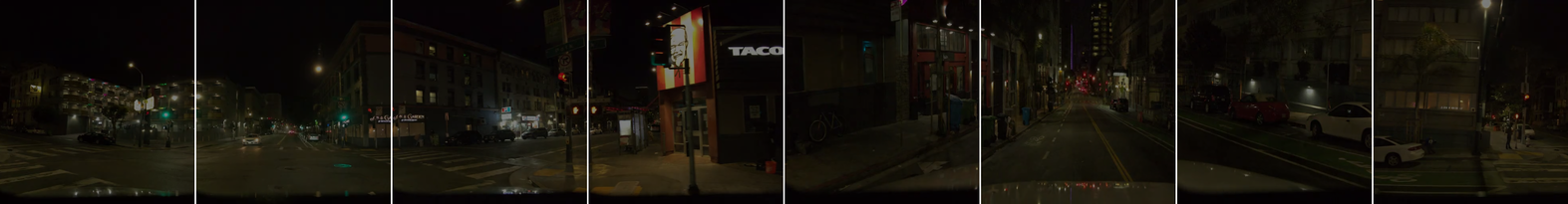}}; 
\node[anchor=east, align=right] at (image8.west) {Target \\ 8:00PM};
\end{tikzpicture}
}

\caption{
\textbf{More Visualizations on time of day editing.} We uniformly sample times between 7 PM and 8 PM to edit the source images (first row), effectively simulating day-to-night transitions.
}
\label{fig:time_edit2}
\end{figure*}

We provide additional visualizations of time-of-day editing results in \figref{time_edit1} and \figref{time_edit2}. By evenly sampling times between 7 PM and 8 PM, we apply all these times to edit the source images. The results demonstrate a seamless transition from day to night, while ensuring 1) geometric consistency between the edited outputs and the original source images, and 2) consistent editing across all results. This highlights the robustness and effectiveness of SceneCrafter.

\begin{figure*}[t]
    \centering

\resizebox{\linewidth}{!}{
\begin{tikzpicture}

\def\imagewidth{14.4cm}
\def\imageheight{1.8cm}
\def\verticalgap{-0.25cm} 

\node (image1) {\includegraphics[width=\imagewidth, height=\imageheight]{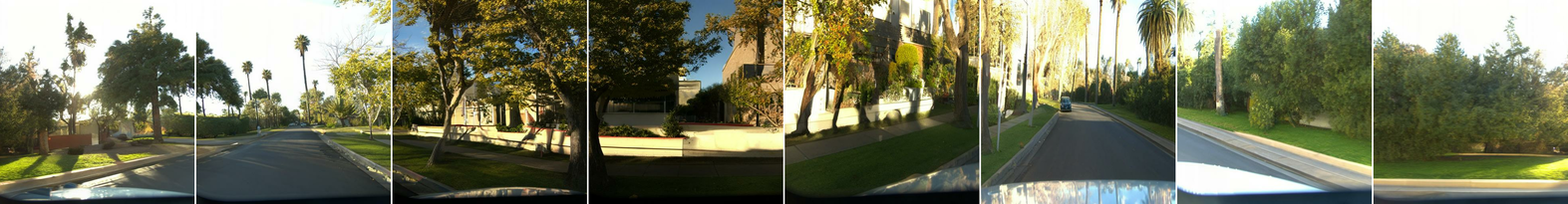}}; 
\node[anchor=east, align=center] at (image1.west) {Source \\ Sunny}; 

\node[below=\verticalgap of image1] (image2) {\includegraphics[width=\imagewidth, height=\imageheight]{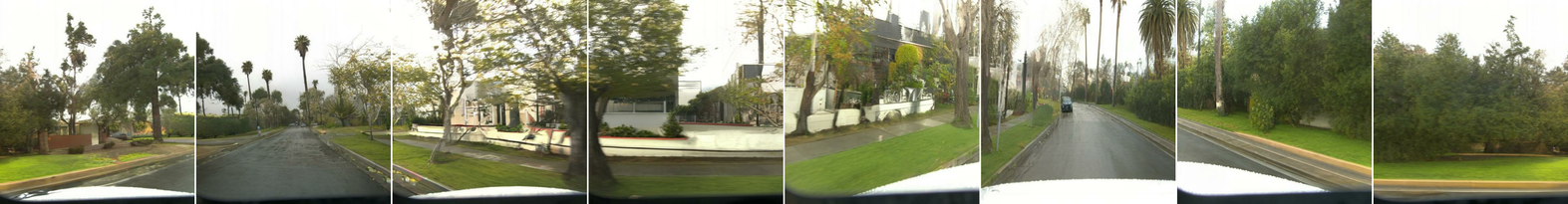}}; 
\node[anchor=east, align=right] at (image2.west) {Target \\ Rainy}; 

\node[below=\verticalgap of image2] (image3) {\includegraphics[width=\imagewidth, height=\imageheight]{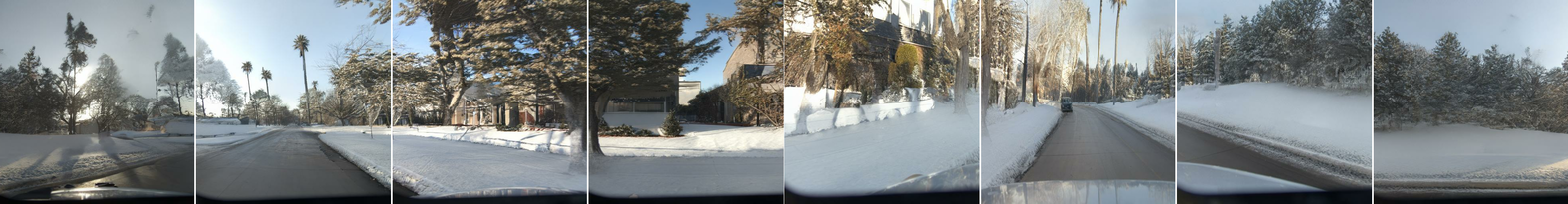}}; 
\node[anchor=east, align=right] at (image3.west) {Target \\ Snowy};

\node[below=\verticalgap of image3] (image4) {\includegraphics[width=\imagewidth, height=\imageheight]{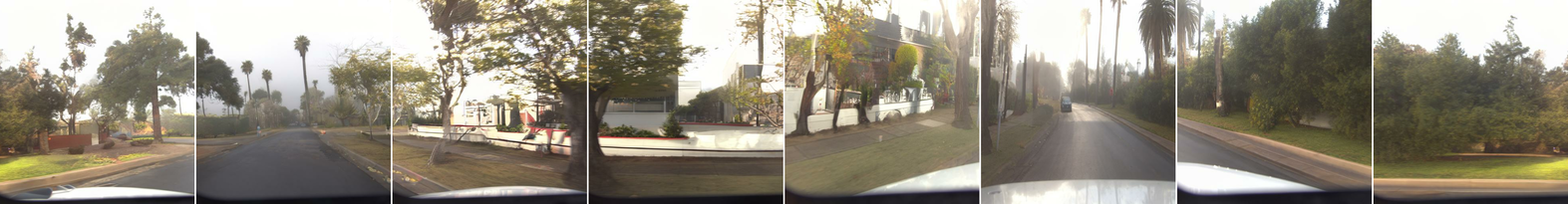}}; 
\node[anchor=east, align=right] at (image4.west) {Target \\ Foggy};

\draw[thick, dashed] ($(image4.south west)+(0,-0.5)$) -- ($(image4.south east)+(0,-0.5)$);

\node[below=1.2cm of image4] (image5) {\includegraphics[width=\imagewidth, height=\imageheight]{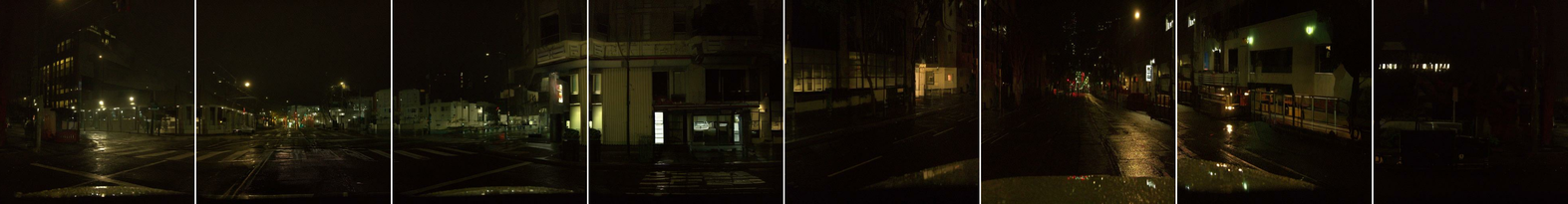}}; 
\node[anchor=east, align=right] at (image5.west) {Source \\ Rainy};

\node[below=\verticalgap of image5] (image6) {\includegraphics[width=\imagewidth, height=\imageheight]{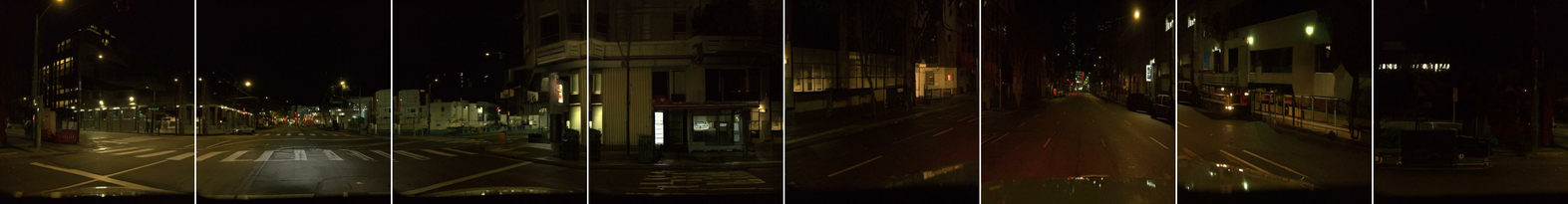}}; 
\node[anchor=east, align=right] at (image6.west) {Target \\ Sunny};

\node[below=\verticalgap of image6] (image7) {\includegraphics[width=\imagewidth, height=\imageheight]{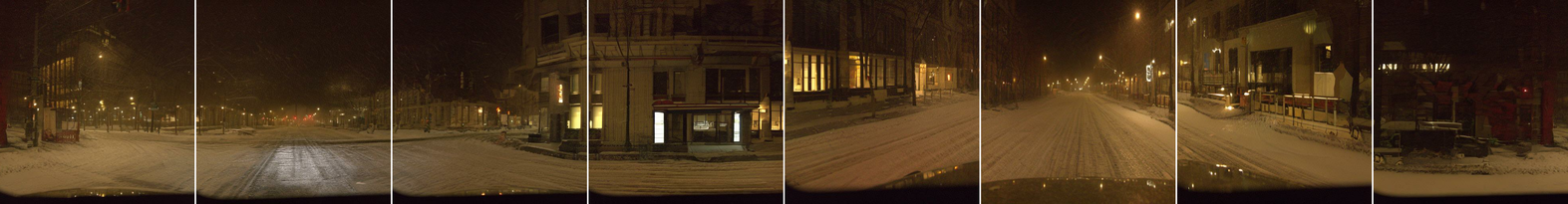}}; 
\node[anchor=east, align=right] at (image7.west) {Target \\ Snowy};

\node[below=\verticalgap of image7] (image8) {\includegraphics[width=\imagewidth, height=\imageheight]{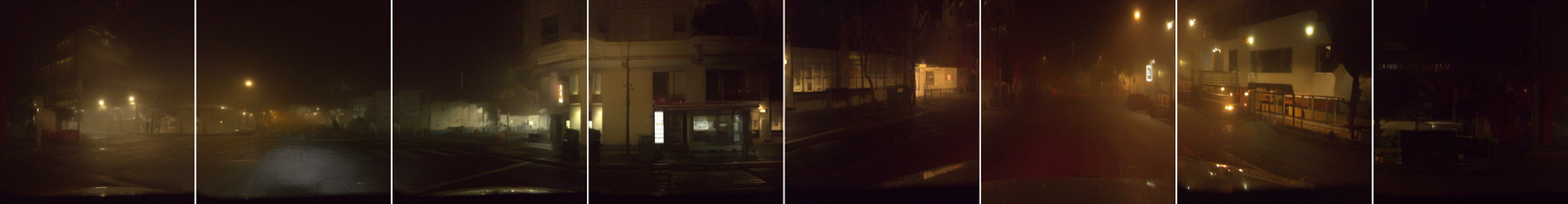}}; 
\node[anchor=east, align=right] at (image8.west) {Target \\ Foggy};

\end{tikzpicture}
}

\caption{
\textbf{More Visualizations on weather editing.} Given images captured under any specific weather (first row), our model transforms the scenes into other weather, including sunny, rainy, snowy, and foggy. The results maintain geometric consistency across all views while reflecting the intended weather effects.
}
\label{fig:weather_edit}
\end{figure*}

\figref{weather_edit} further showcases the weather editing results. Starting with multi-view images captured under a specific weather condition, we edit them to different weather conditions, including sunny, rainy, foggy, and snowy. Even when tackling challenging edits like snowy weather, which significantly differs from the original, SceneCrafter effectively preserves the scene's original layout like trees and roads, while seamlessly adding realistic elements like snow covering roads or grass. This makes SceneCrafter a practical choice for simulating extreme weather conditions.

\begin{figure*}[t]
    \centering

\resizebox{\linewidth}{!}{
\begin{tikzpicture}

\def\imagewidth{14.4cm}
\def\imageheight{1.8cm}
\def\verticalgap{-0.25cm} 

\node (image1) {\includegraphics[width=\imagewidth, height=\imageheight]{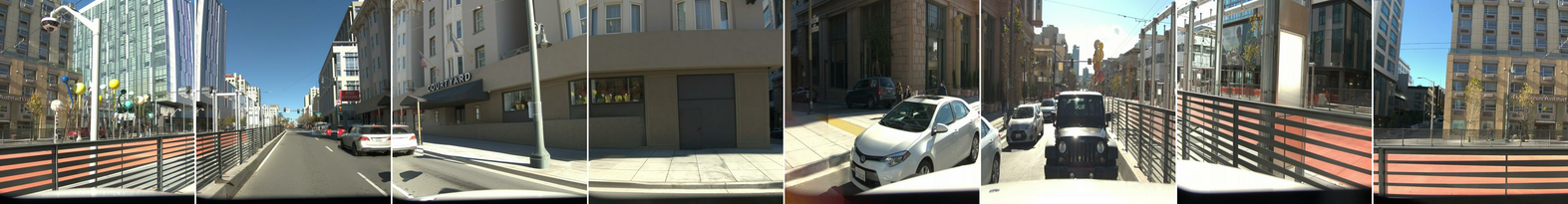}}; 
\node[anchor=east, align=center] at (image1.west) {Source}; 

\node[below=\verticalgap of image1] (image2) {\includegraphics[width=\imagewidth, height=\imageheight]{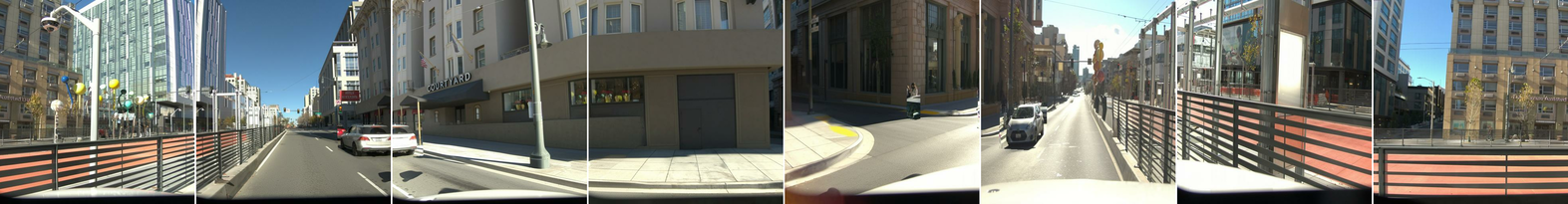}}; 
\node[anchor=east, align=right] at (image2.west) {Target \\ Example1}; 

\node[below=\verticalgap of image2] (image3) {\includegraphics[width=\imagewidth, height=\imageheight]{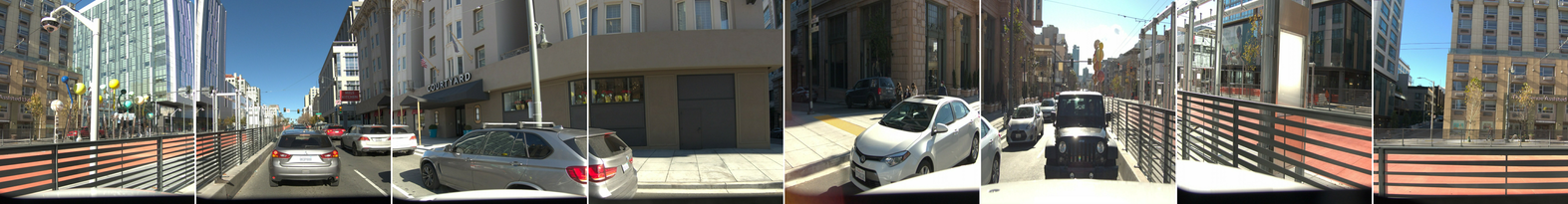}}; 
\node[anchor=east, align=right] at (image3.west) {Target \\ Example2};

\node[below=\verticalgap of image3] (image4) {\includegraphics[width=\imagewidth, height=\imageheight]{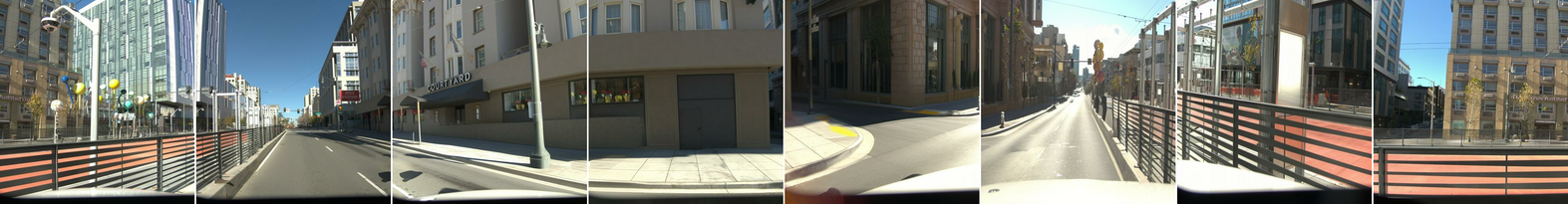}}; 
\node[anchor=east, align=right] at (image4.west) {Target \\ Example3};

\draw[thick, dashed] ($(image4.south west)+(0,-0.5)$) -- ($(image4.south east)+(0,-0.5)$);

\node[below=1.2cm of image4] (image5) {\includegraphics[width=\imagewidth, height=\imageheight]{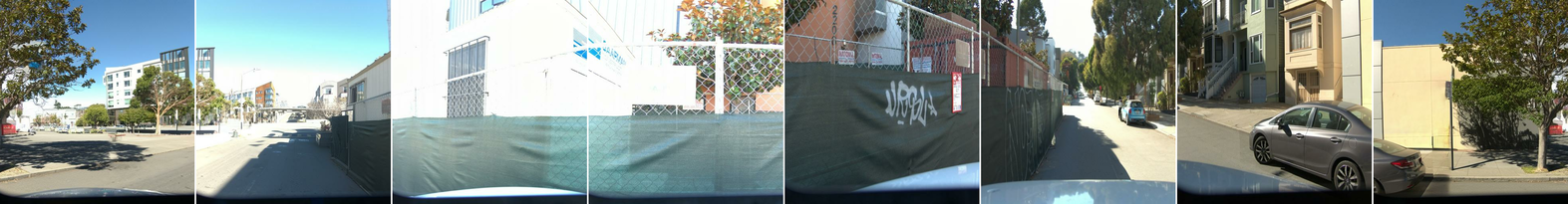}}; 
\node[anchor=east, align=right] at (image5.west) {Source};

\node[below=\verticalgap of image5] (image6) {\includegraphics[width=\imagewidth, height=\imageheight]{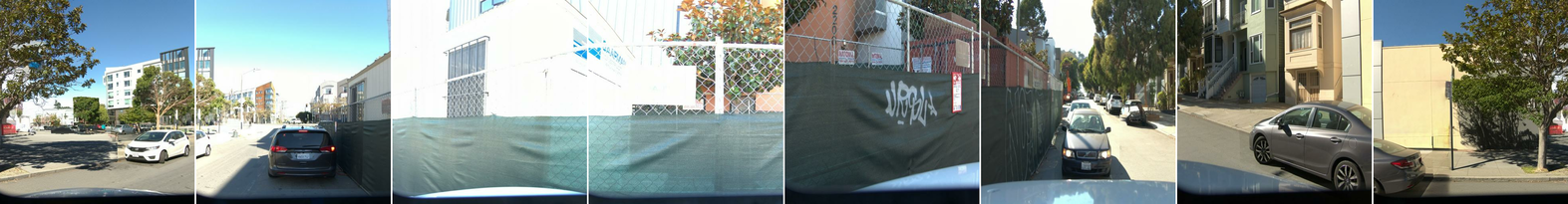}}; 
\node[anchor=east, align=right] at (image6.west) {Target \\ Example1};

\node[below=\verticalgap of image6] (image7) {\includegraphics[width=\imagewidth, height=\imageheight]{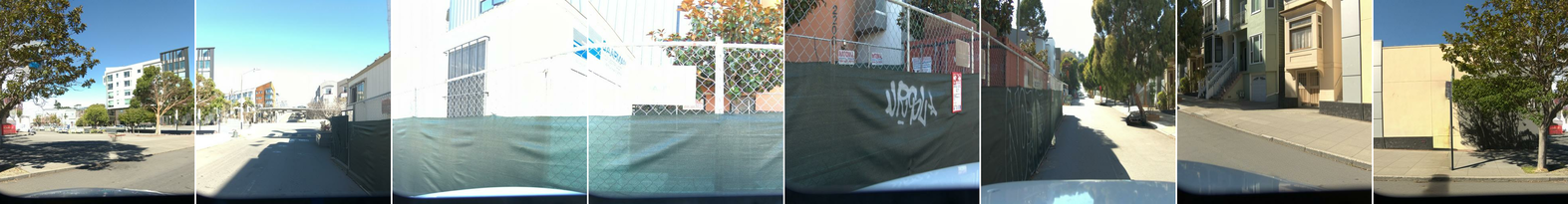}}; 
\node[anchor=east, align=right] at (image7.west) {Target \\ Example2};

\node[below=\verticalgap of image7] (image8) {\includegraphics[width=\imagewidth, height=\imageheight]{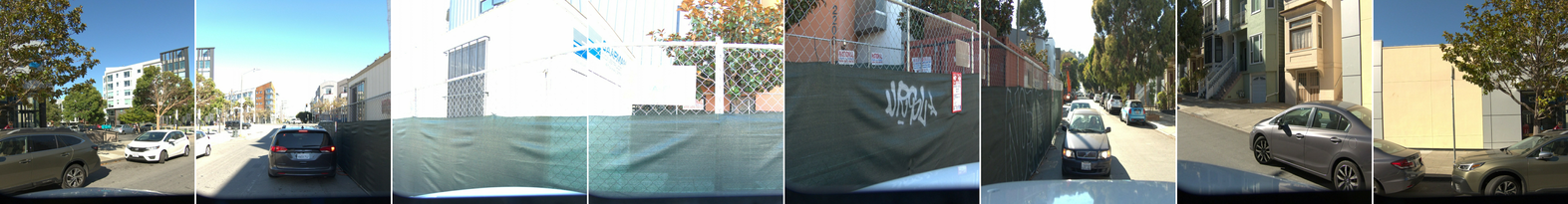}}; 
\node[anchor=east, align=right] at (image8.west) {Target \\ Example3};

\end{tikzpicture}
}

\caption{
\textbf{More Visualizations on local editing.} SceneCrafter enables the insertion or removal of arbitrary agents within the source images. We demonstrate three editing examples conditioned on different agent boxes. SceneCrafter exhibits strong robustness across diverse agent box conditions, generating highly realistic vehicles or inpainted backgrounds.
}
\label{fig:inpaint_edit1}
\end{figure*}

\figref{inpaint_edit1} illustrates the agent editing results. SceneCrafter allows for insertion or removal of arbitrary agents in the source images, controlled through specified agent boxes and their types (foreground for insertion and background for removal). The results demonstrate SceneCrafter's ability to perform precise and dexterous scene manipulations, even for small vehicles. Additionally, the generated vehicles are highly realistic, appearing nearly indistinguishable from real ones and aligning perfectly with the lighting conditions of the original scenes.

\begin{figure*}[t]
    \centering

\resizebox{\linewidth}{!}{
\begin{tikzpicture}

\def\imagewidth{14.4cm}
\def\imageheight{1.8cm}
\def\verticalgap{-0.25cm} 

\node (image1) {\includegraphics[width=\imagewidth, height=\imageheight]{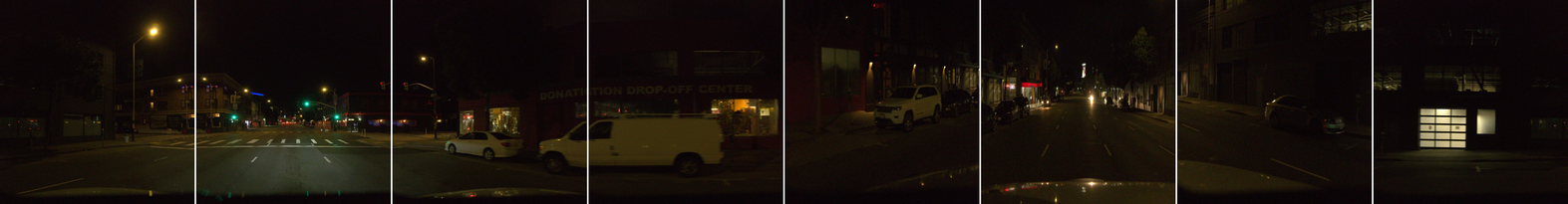}}; 
\node[anchor=east, align=center] at (image1.west) {Source}; 

\node[below=\verticalgap of image1] (image2) {\includegraphics[width=\imagewidth, height=\imageheight]{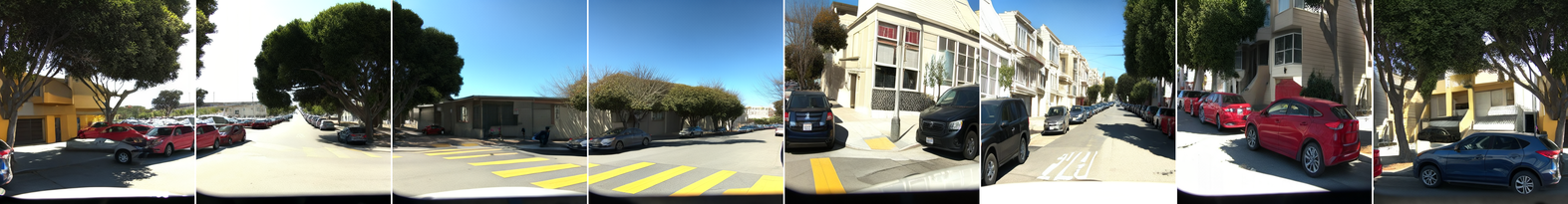}}; 
\node[anchor=east, align=right] at (image2.west) {Target \\ SDEdit}; 

\node[below=\verticalgap of image2] (image3) {\includegraphics[width=\imagewidth, height=\imageheight]{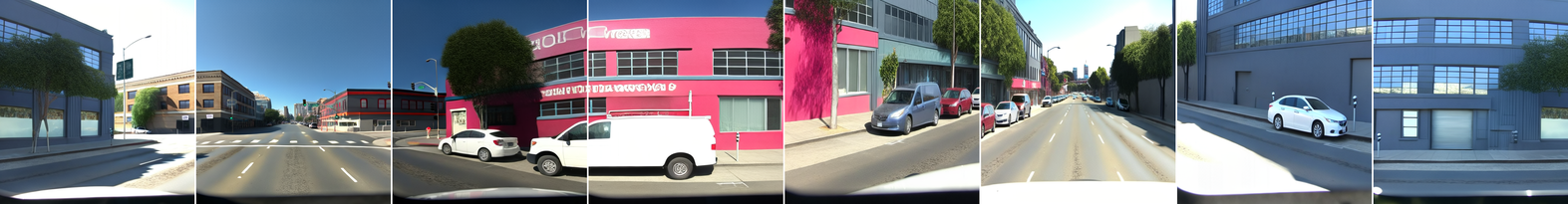}}; 
\node[anchor=east, align=right] at (image3.west) {Target \\ P2P*};

\node[below=\verticalgap of image3] (image4) {\includegraphics[width=\imagewidth, height=\imageheight]{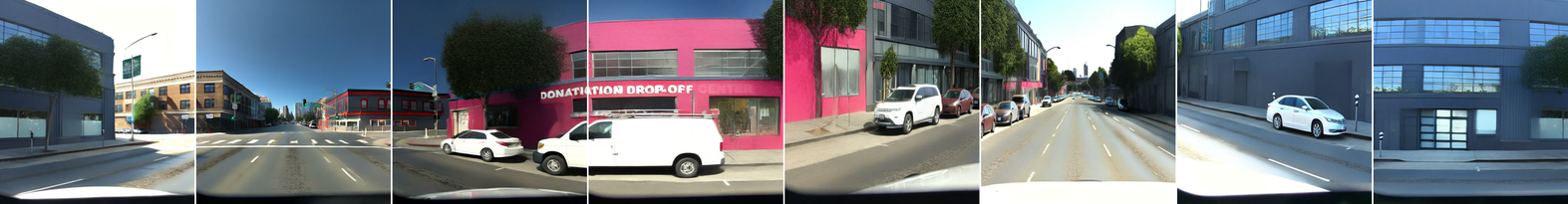}}; 
\node[anchor=east, align=right] at (image4.west) {Target \\ Ours};

\draw[thick, dashed] ($(image4.south west)+(0,-0.5)$) -- ($(image4.south east)+(0,-0.5)$);

\node[below=1.2cm of image4] (image5) {\includegraphics[width=\imagewidth, height=\imageheight]{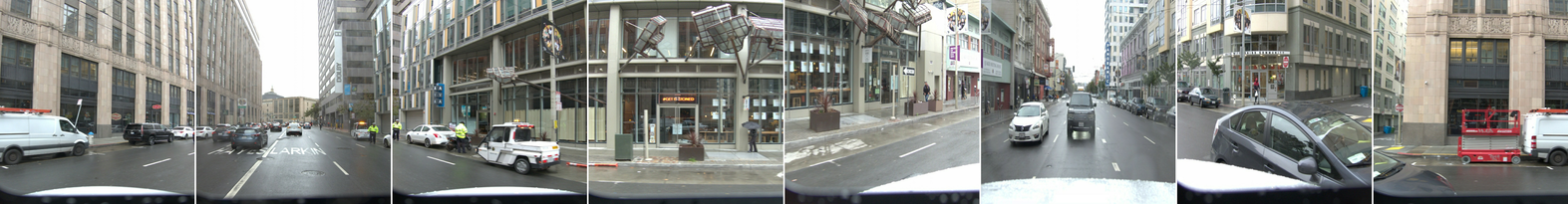}}; 
\node[anchor=east, align=right] at (image5.west) {Source};

\node[below=\verticalgap of image5] (image6) {\includegraphics[width=\imagewidth, height=\imageheight]{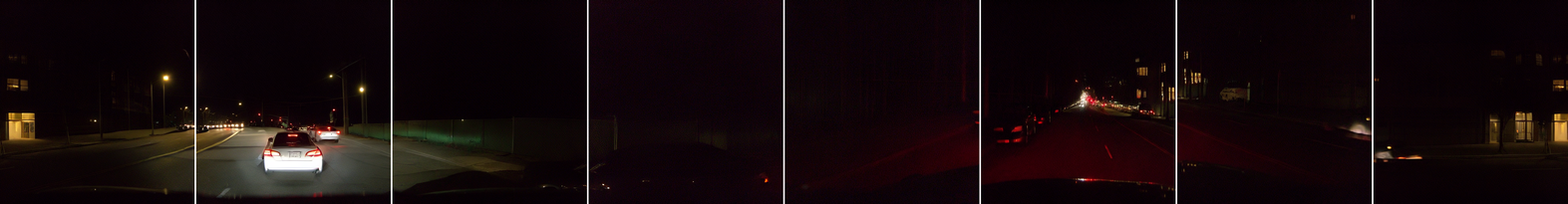}}; 
\node[anchor=east, align=right] at (image6.west) {Target \\ SDEdit};

\node[below=\verticalgap of image6] (image7) {\includegraphics[width=\imagewidth, height=\imageheight]{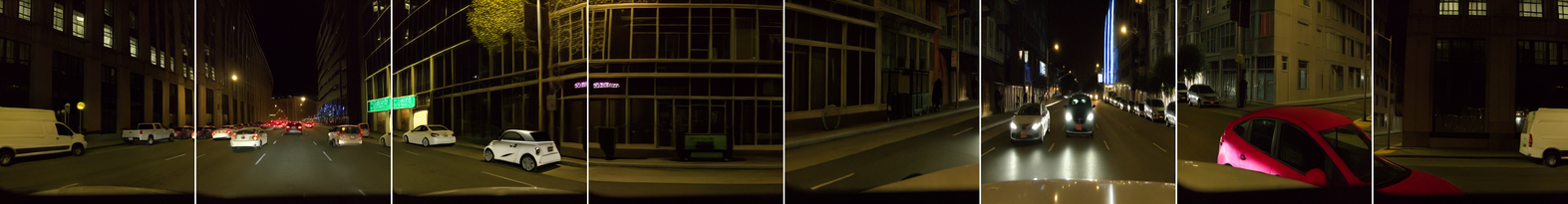}}; 
\node[anchor=east, align=right] at (image7.west) {Target \\ P2P*};

\node[below=\verticalgap of image7] (image8) {\includegraphics[width=\imagewidth, height=\imageheight]{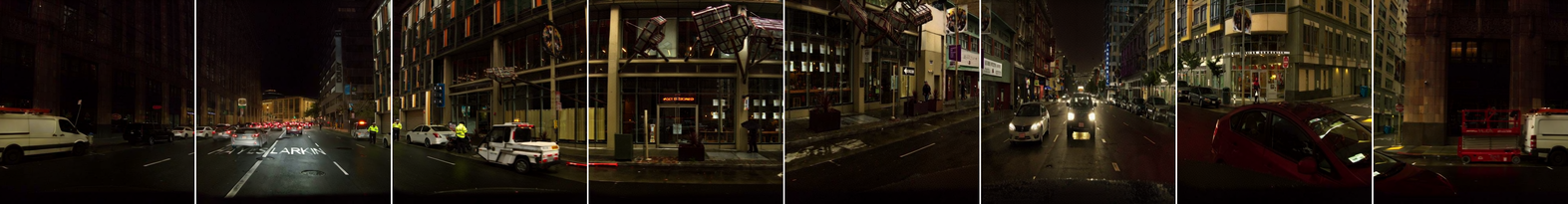}}; 
\node[anchor=east, align=right] at (image8.west) {Target \\ Ours};

\end{tikzpicture}
}

\caption{
\textbf{Qualitative comparison with SDEdit and P2P* baselines on time of day editing.} These scenes are included as part of our user study data. In the user study, 9 out of 11 participants rated SceneCrafter as having the best editing results for the first scene, with 2 preferred P2P*. For the second scene, 10 out of 11 participants favored SceneCrafter and 1 preferred P2P*.
}
\label{fig:time_compare}
\end{figure*}

\begin{figure*}[t]
    \centering

\resizebox{\linewidth}{!}{
\begin{tikzpicture}

\def\imagewidth{14.4cm}
\def\imageheight{1.8cm}
\def\verticalgap{-0.25cm} 

\node (image1) {\includegraphics[width=\imagewidth, height=\imageheight]{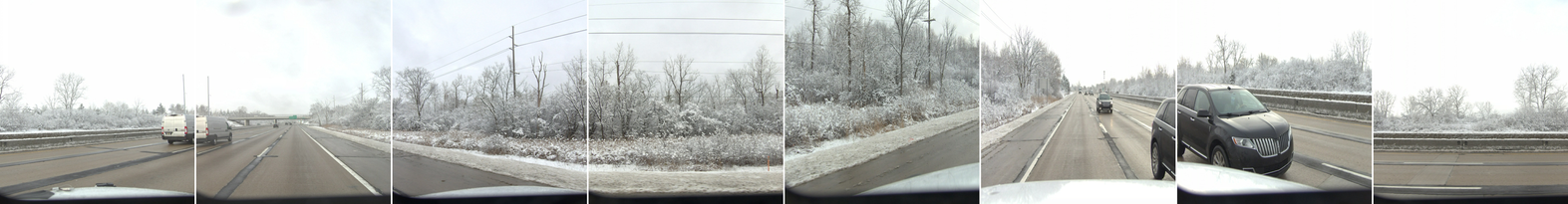}}; 
\node[anchor=east, align=center] at (image1.west) {Source}; 

\node[below=\verticalgap of image1] (image2) {\includegraphics[width=\imagewidth, height=\imageheight]{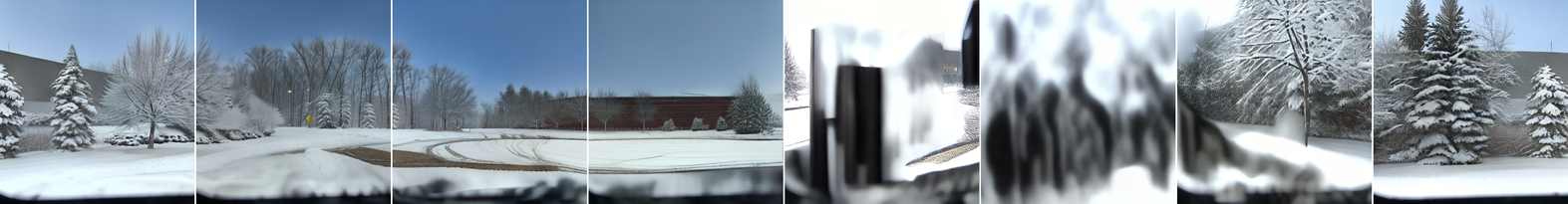}}; 
\node[anchor=east, align=right] at (image2.west) {Target \\ SDEdit}; 

\node[below=\verticalgap of image2] (image3) {\includegraphics[width=\imagewidth, height=\imageheight]{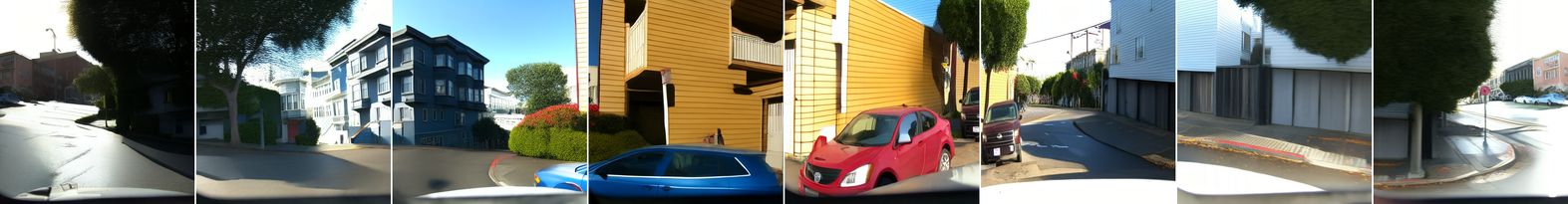}}; 
\node[anchor=east, align=right] at (image3.west) {Target \\ P2P*};

\node[below=\verticalgap of image3] (image4) {\includegraphics[width=\imagewidth, height=\imageheight]{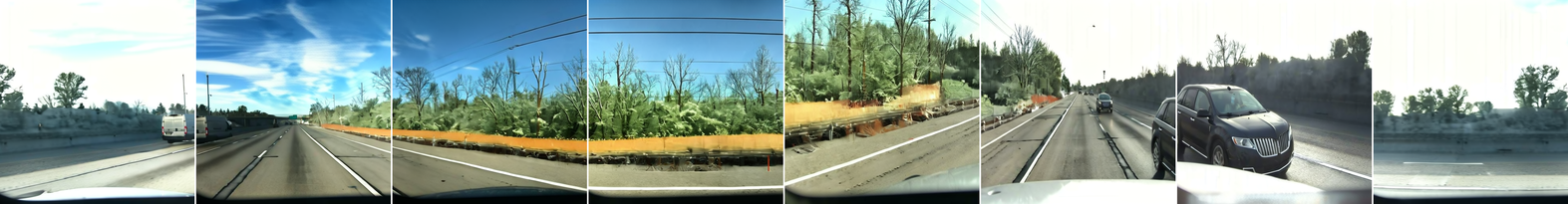}}; 
\node[anchor=east, align=right] at (image4.west) {Target \\ Ours};

\draw[thick, dashed] ($(image4.south west)+(0,-0.5)$) -- ($(image4.south east)+(0,-0.5)$);

\node[below=1.2cm of image4] (image5) {\includegraphics[width=\imagewidth, height=\imageheight]{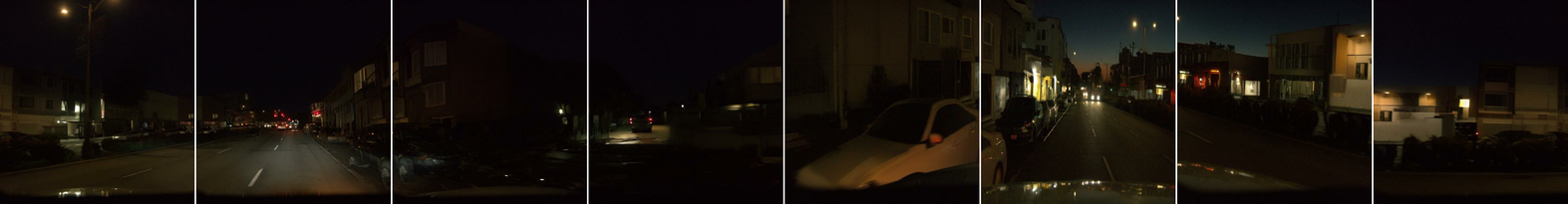}}; 
\node[anchor=east, align=right] at (image5.west) {Source};

\node[below=\verticalgap of image5] (image6) {\includegraphics[width=\imagewidth, height=\imageheight]{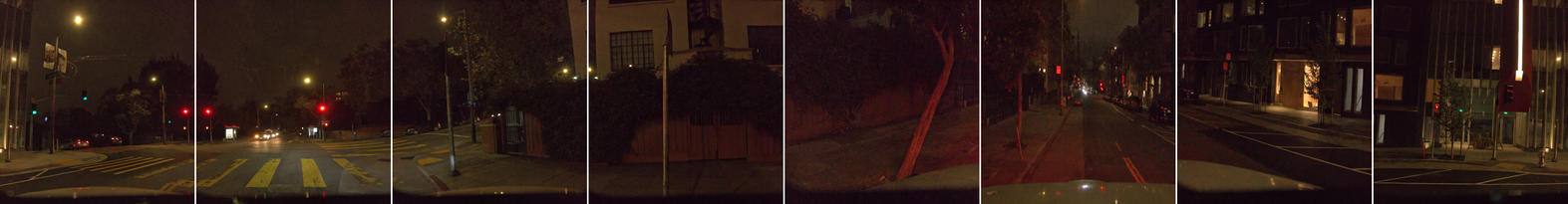}}; 
\node[anchor=east, align=right] at (image6.west) {Target \\ SDEdit};

\node[below=\verticalgap of image6] (image7) {\includegraphics[width=\imagewidth, height=\imageheight]{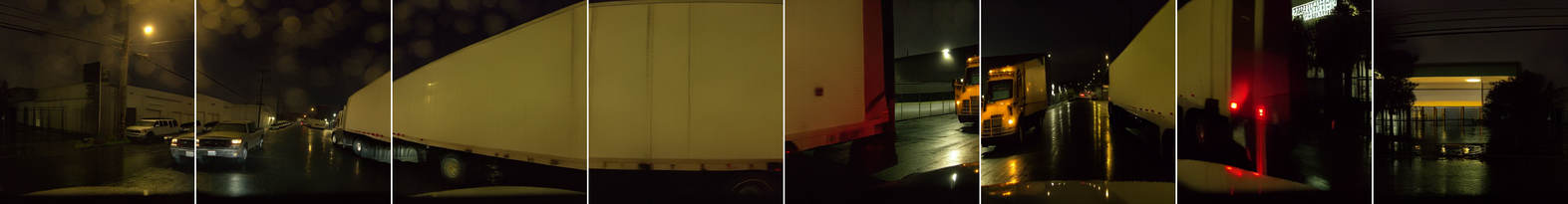}}; 
\node[anchor=east, align=right] at (image7.west) {Target \\ P2P*};

\node[below=\verticalgap of image7] (image8) {\includegraphics[width=\imagewidth, height=\imageheight]{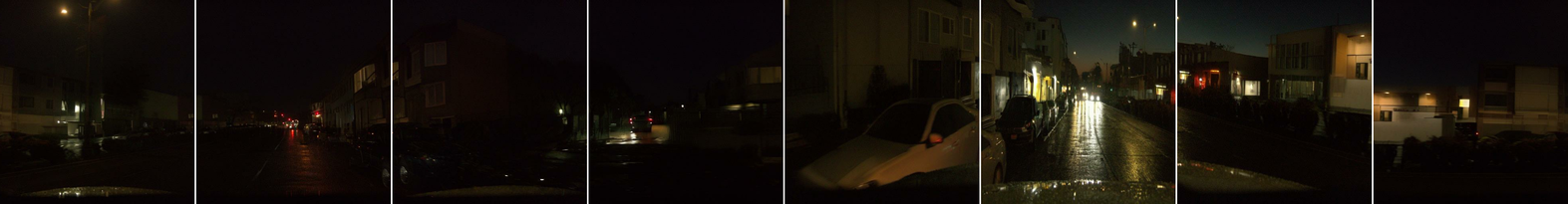}}; 
\node[anchor=east, align=right] at (image8.west) {Target \\ Ours};

\end{tikzpicture}
}

\caption{
\textbf{Qualitative comparison with SDEdit and P2P* baselines on weather editing.} In the user study, all participants preferred the editing results generated by our method for both scenes, demonstrating its strong alignment with human preferences.
}
\label{fig:weather_compare}
\end{figure*}

We present more comparison results for global edits with SDEdit~\cite{meng2021sdedit} and Prompt-to-Prompt~\cite{hertz2022prompt} in \figref{time_compare} and \figref{weather_compare}. Our SceneCrafter excels at preserving fine-grained details from the source images, such as text and icons, while delivering realistic and precise edits. These scenes are also included as part of our user study data. For the first example of time-of-day editing, 9 out of 11 users rated SceneCrafter as having the best editing results, while 2 preferred P2P*. For the first example of weather editing, all participants unanimously selected SceneCrafter as the best performer, showing its strong alignment with human preferences.

\begin{figure*}[t]
    \centering

\resizebox{\linewidth}{!}{
\begin{tikzpicture}

\def\imagewidth{14.4cm}
\def\imageheight{1.8cm}
\def\verticalgap{-0.25cm} 

\node (image1) {\includegraphics[width=\imagewidth, height=\imageheight]{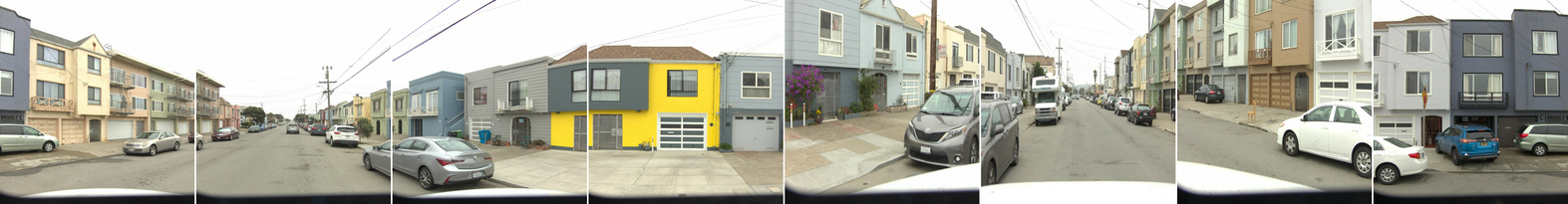}}; 
\node[anchor=east, align=center] at (image1.west) {Source}; 

\node[below=\verticalgap of image1] (image2) {\includegraphics[width=\imagewidth, height=\imageheight]{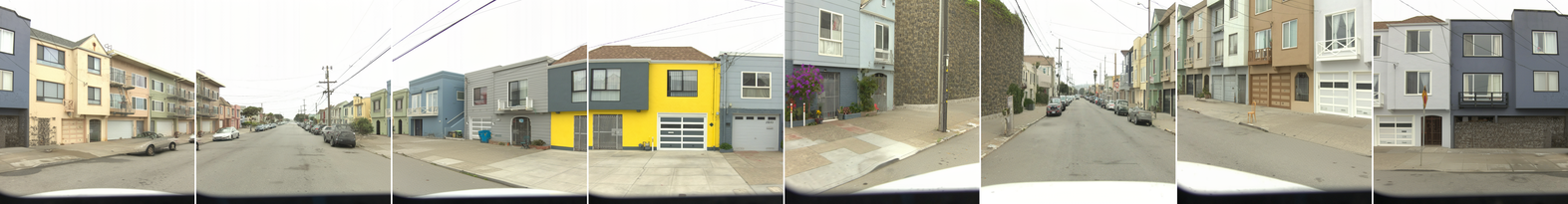}}; 
\node[anchor=east, align=right] at (image2.west) {Target \\ 2D-Repaint}; 

\node[below=\verticalgap of image2] (image3) {\includegraphics[width=\imagewidth, height=\imageheight]{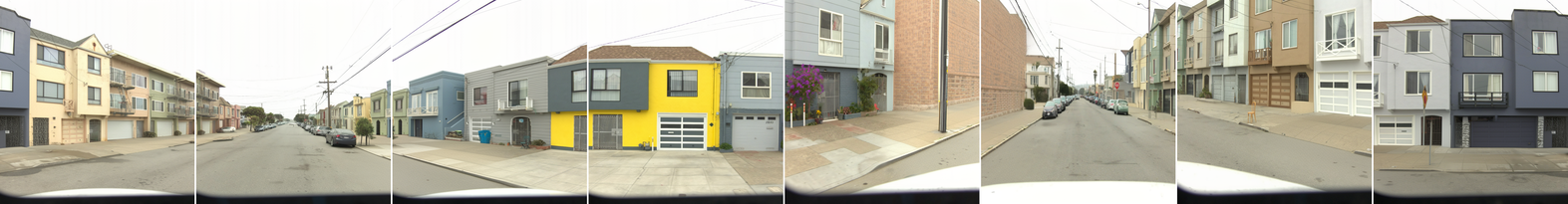}}; 
\node[anchor=east, align=right] at (image3.west) {Target \\ MV-Repaint};

\node[below=\verticalgap of image3] (image4) {\includegraphics[width=\imagewidth, height=\imageheight]{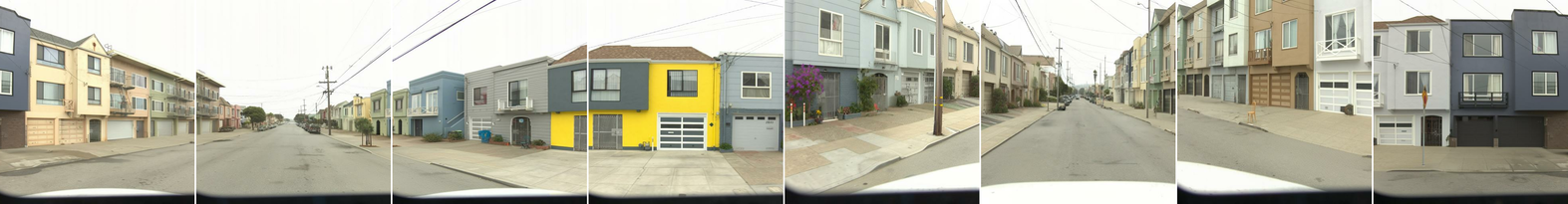}}; 
\node[anchor=east, align=right] at (image4.west) {Target \\ Ours};

\draw[thick, dashed] ($(image4.south west)+(0,-0.5)$) -- ($(image4.south east)+(0,-0.5)$);

\node[below=1.2cm of image4] (image5) {\includegraphics[width=\imagewidth, height=\imageheight]{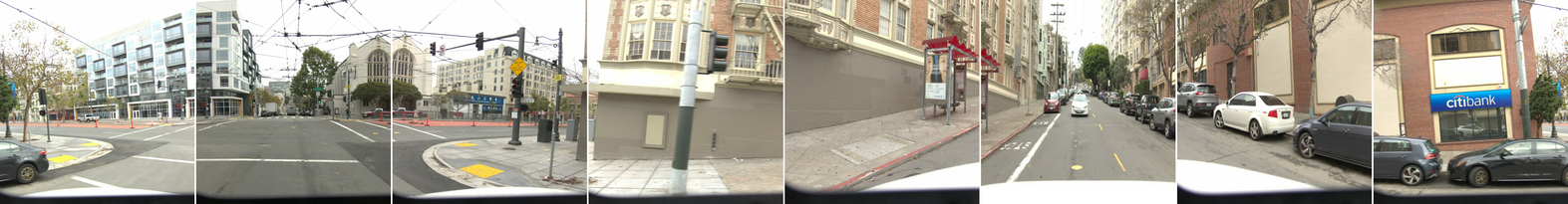}}; 
\node[anchor=east, align=right] at (image5.west) {Source};

\node[below=\verticalgap of image5] (image6) {\includegraphics[width=\imagewidth, height=\imageheight]{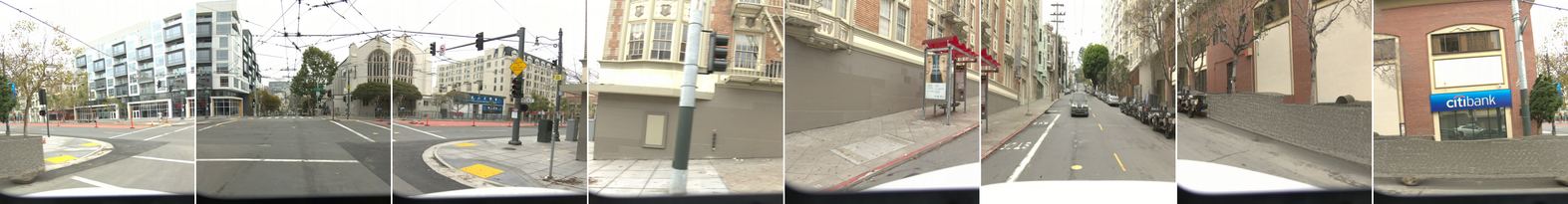}}; 
\node[anchor=east, align=right] at (image6.west) {Target \\ 2D-Repaint};

\node[below=\verticalgap of image6] (image7) {\includegraphics[width=\imagewidth, height=\imageheight]{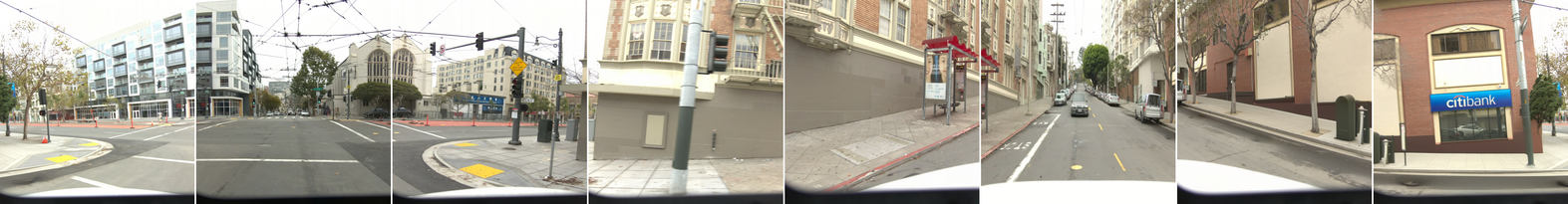}}; 
\node[anchor=east, align=right] at (image7.west) {Target \\ MV-Repaint};

\node[below=\verticalgap of image7] (image8) {\includegraphics[width=\imagewidth, height=\imageheight]{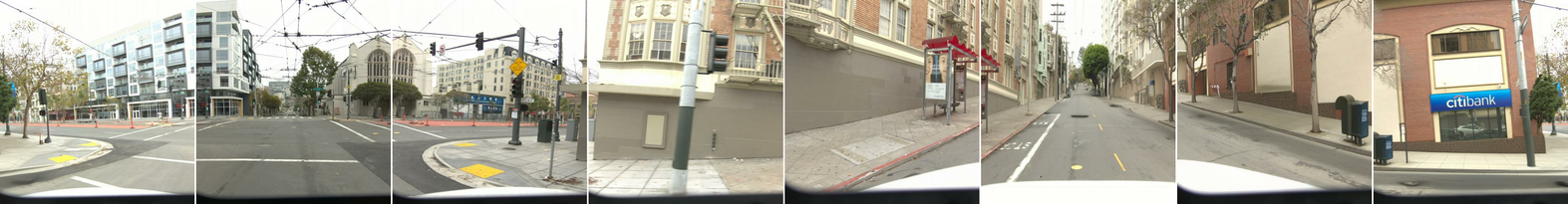}}; 
\node[anchor=east, align=right] at (image8.west) {Target \\ Ours};

\end{tikzpicture}
}

\caption{
\textbf{Qualitative comparison with 2D-Repaint and MV-Repaint baselines on local editing.} Our method consistently achieves superior results in completely removing all vehicles from the scenes.
}
\label{fig:inpaint_compare}
\end{figure*}
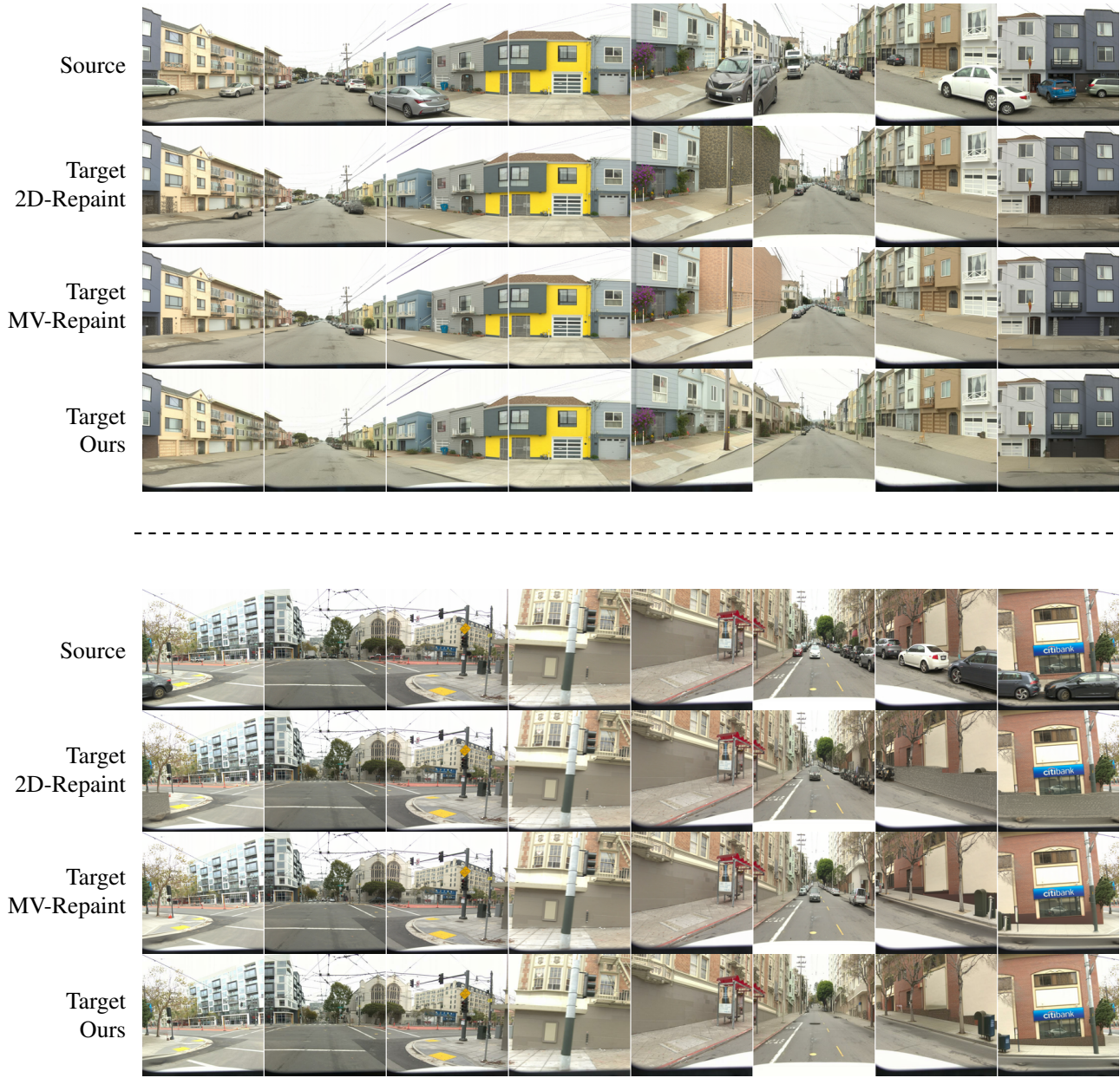

\figref{inpaint_compare} offers a visual comparison of vehicle removal capabilities between our method and the 2D-Repaint and MV-Repaint baselines.  2D-Repaint relies on Stable Diffusion, and struggles to maintain view consistency in 3D editing task. While MV-Repaint leverages global editing teacher model, it lacks masked training priors, resulting in incomplete car removals. In contrast, our method, trained on alpha-blended pairs of "empty streets" and "populated streets," demonstrates superior capability in handling complex scene editing with precision and consistency.

\clearpage
\clearpage

\begin{figure*}[htbp]
    \centering
    \includegraphics[width=1.7\columnwidth]{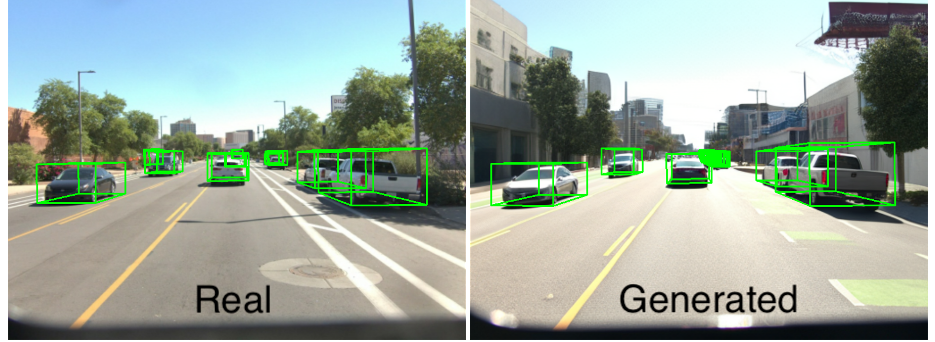}
    \caption{Comparing detection results on real logs and synthetic images generated by SceneCrafter.}
    \label{fig:detection}
\end{figure*}

\begin{figure*}[htbp]
    \centering
    \includegraphics[width=1.7\columnwidth]{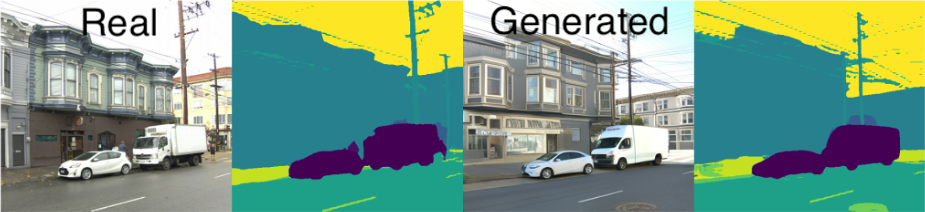}
    \caption{Comparing segmentation results on real logs and synthetic images generated by SceneCrafter.}
    \label{fig:segmentation}
\end{figure*}

\section{Downstream Task Performance}
The theme for this project is on simulation, specifically evaluating downstream models within a controllable simulated environment.
Thus, we test off-the-shelf downstream models on our generated data without retraining them.

\subsection{Segmentation}
We applied a Panoptic-DeepLab semantic segmentation model on real and conditionally generated camera images, and found its behavior to be similar. The KL divergence of aggregate class distributions of 26 classes across 8,000 images is only 0.01133, with the only minor variations being in vegetation, building and sky due to SceneCrafter being least conditioned on their outline. The predicted masks are sharp and semantically corresponding.

\subsection{Detection}
We employed a monocular 3D detection model to compare predictions between real and conditionally generated front camera images. We find that the 3D detector holds up surprisingly well to the generated images. It is rare to find false-negative detections in the generated imagery.

\end{document}